\documentclass{article}

\usepackage[main, final]{neurips_2026}

\usepackage[utf8]{inputenc}
\usepackage[T1]{fontenc}
\usepackage{url}
\usepackage{booktabs}
\usepackage{amsfonts}
\usepackage{nicefrac}
\usepackage{microtype}
\usepackage{xcolor}
\usepackage{amsmath}
\usepackage{graphicx}
\usepackage{tcolorbox}
\tcbuselibrary{breakable}
\usepackage{multirow}
\usepackage{enumitem}
\usepackage{algorithm}
\usepackage{algpseudocode}
\usepackage{upgreek}
\usepackage[colorlinks=true]{hyperref}

\usepackage{marvosym}

\DeclareMathOperator*{\argmax}{arg\,max}

\title{Failure Profiling and Reachable Trajectory Selection for Reasoning Distillation}

\author{Yijun Zhu\textsuperscript{1} \quad Jianxin Wang\textsuperscript{1} \quad Chengchao Shen\textsuperscript{1, \Letter}\\
  \textsuperscript{1}School of Computer Science and Engineering, Central South University \\
  \texttt{zhuyij@csu.edu.cn, jxwang@mail.csu.edu.cn, scc.cs@csu.edu.cn}
}

\begin{document}

\maketitle

\raggedbottom
\renewcommand{\thefootnote}{\Letter}
\footnotetext[1]{Corresponding author.}
\renewcommand{\thefootnote}{\arabic{footnote}}
\setcounter{footnote}{0}

\begin{abstract}
Training smaller language models on reasoning solutions generated by a larger teacher model is the prevailing approach for transferring mathematical reasoning capabilities.
However, the student tends to overfit the output distribution of the teacher and generalizes poorly when its own behavior at inference time deviates from the training data.
Recent methods attempt to bridge this gap by incorporating error data, yet the corrective trajectories still reflect the reasoning patterns of the teacher, which the student cannot reliably reproduce, leaving the distribution mismatch largely unresolved.
We propose a method, Failure Profiling and Reachable Trajectory Selection (FPRT), built on the insight that effective training data should be jointly determined by what the student fails on and what it can realistically learn.
Our method first profiles the failure behavior of the student on each problem through repeated sampling, characterizing it in terms of error rate and error consistency to reveal qualitatively distinct failure modes.
Each mode demands a fundamentally different training signal.
The diagnosed mode then guides candidate generation from the teacher via distinct prompting strategies tailored to each failure type.
Among the correct candidates, our method selects the one whose initial reasoning steps after the point of divergence are most reproducible by the student.
These first corrective steps constitute the primary bottleneck, while the remaining steps follow an already corrected context and can therefore be continued more readily.
This contrasts with standard reranking approaches, which score the trajectory as a whole and thereby dilute the signal at the critical reasoning transition.
Experiments on GSM8K, MATH, and three out-of-distribution benchmarks for Llama-3.1-8B and DeepSeek-Math-7B demonstrate that our method achieves the SOTA performance across the five benchmarks on both student backbones.
The source code and trained weights are available at \url{https://github.com/visresearch/FPRT}.
\end{abstract}

\section{Introduction}\label{sec:intro}

\begin{figure}[t]
\centering
\includegraphics[width=\linewidth]{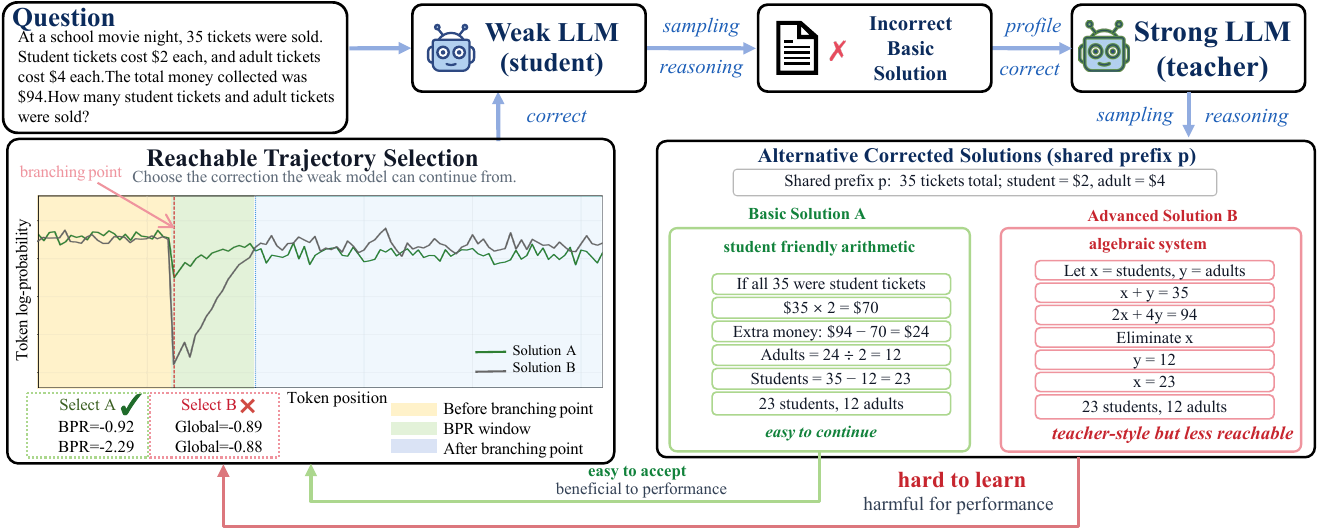}
\vspace{-1.5em}
\caption{
Due to the capability gap between a strong teacher and a weaker student, teacher-generated reasoning trajectories may be difficult for the student to reproduce and may hinder learning. We guide the teacher to generate correction candidates conditioned on the student's errors and select the candidate with the highest branching point reachability score. This score averages student token log-probabilities within a local window immediately after the first divergence, reducing dilution from later tokens.
}
\vspace{-0.8em}
\label{fig:motivation}
\end{figure}

Frontier large language models (LLMs), such as
GPT-4~\cite{achiam2023gpt},
DeepSeek-R1~\cite{guo2025deepseek}, and
Qwen2.5~\cite{yang2024qwen2}, have demonstrated strong
mathematical reasoning ability. In contrast, smaller models
such as Llama-3.1-8B~\cite{dubey2024llama} and
DeepSeek-Math-7B~\cite{shao2024deepseekmath} still struggle
with complex mathematical problems. A growing line of
research seeks to bridge this gap by training smaller
students on reasoning traces generated by more capable
teachers~\cite{magister2023teaching, ho2023large,
li2024common, yu2024metamath}. Despite substantial
progress, the resulting student overfits the output
distribution of the teacher: trained exclusively on correct
demonstrations that mirror the reasoning style of the
teacher, the student has no reliable way to recover once
its behavior at inference time deviates from that
distribution. The cost is especially high in mathematical
reasoning, where each step depends on all preceding ones
and even a single misstep invalidates the entire subsequent
chain.

Recent work has begun to incorporate error data into
training to address this gap.
LEMMA~\cite{pan2025lemma} pairs erroneous solutions with
corrective reflections.
Self-Correct~\cite{welleck2023generating} explores minimal
edit corrections that fix mistakes generated by the model.
Approaches based on MCTS systematically explore error nodes
in the reasoning tree~\cite{xie2024montecarlo,
chen2024alphamath}.
These methods mark meaningful progress, yet they share two limitations.
First, they do not condition on the behavioral failure mode of the student but instead apply the same correction strategy regardless of whether the model is confidently locked onto a wrong answer or producing a different error each time.
Specifically, consistently arriving at the same wrong answer reflects a strong but misguided reasoning pattern, whereas producing diverse errors signals the absence of any stable strategy.
For the above failure modes, a single correction strategy cannot efficiently fix student errors.
Second, even when the failure mode is correctly identified, the corrective trajectories still come from a teacher and
exhibit reasoning patterns that the student cannot reliably reproduce, a problem that prior work on knowledge transfer
has shown to worsen as the capacity gap between teacher and student grows~\cite{mirzadeh2020improved}.
Such corrections risk reducing learning to superficial pattern matching rather than genuine skill acquisition.

To address the above limitations, we introduce a novel failure profiling and reachable trajectory selection framework for reasoning distillation.
In the failure profiling stage, our method characterizes student errors in terms of error rate and error consistency from multiple sampled responses on each problem to identify the underlying failure mode.
Then, we elicit targeted correction candidates from an advanced teacher using prompts tailored to each diagnosed mode.
When the student is locked onto a single wrong answer, the prompt directs the teacher to construct corrections that break the specific misconception, whereas when the student produces diverse errors, the prompt instead requests structured step by step reasoning to supply the missing problem solving framework.

In the reachable trajectory selection stage, we select the most reproducible candidate trajectory produced by the teacher to correct the student's erroneous reasoning steps, thereby improving the student's reasoning ability.
Unlike standard reranking approaches~\cite{zhang2024grape} that score the trajectory as a whole, our method evaluates compatibility within a local window immediately after the position where the reasoning of the student first diverged, as shown in Figure~\ref{fig:motivation}.
We find that the above local window scoring can clearly distinguish the student-friendly correction from other candidates, where later tokens conditioned on a corrected context will not dilute the signal at the critical reasoning transition.
Once the student successfully executes the transition away from its error, the remaining steps follow an already corrected context and can therefore be continued more readily.

Overall, our contributions can be summarized as follows:
\begin{enumerate}
    \item We introduce behavioral profiling based on error
    rate and error consistency to classify model
    errors into distinct failure modes and enable targeted
    correction strategies for qualitatively different error
    patterns.

    \item We propose branching point reachability scoring,
    a trajectory selection mechanism that evaluates the
    log-probability under the student within a local window
    immediately following the point of divergence rather
    than over the full response, thereby guiding the student to focus on the critical reasoning transition where the student makes mistakes.

    \item Extensive experiments on Llama-3.1-8B and
    DeepSeek-Math-7B across the mathematical reasoning
    benchmarks GSM8K and MATH, together with the
    out-of-distribution evaluation sets SVAMP, ASDiv, and
    MAWPS, demonstrate that our method achieves the highest average accuracy across the five benchmarks on both student backbones.
\end{enumerate}

\section{Related Work}
\label{sec:related_work}

\paragraph{Training on Teacher-Generated Reasoning Traces.}
A dominant approach to improving mathematical reasoning in
smaller models is fine-tuning on reasoning traces from
more capable
models~\cite{yu2024metamath,luo2023wizardmath,yue2024mammoth,yue2024mammoth2,lu2024mathgenie,li2024common}.
Rejection sampling fine-tuning~\cite{yuan2023scaling}
trains on correct paths sampled from the student, and
DART-Math~\cite{tong2024dartmath} rebalances this corpus
toward harder queries. LLM2LLM~\cite{lee2024llm2llm}
iteratively synthesizes new data targeting problems on
which the student fails. A separate line tackles the
distribution mismatch between teacher data and student
capability. GKD~\cite{agarwal2024onpolicy} trains the
student on self-generated sequences with teacher feedback,
and Speculative Knowledge
Distillation~\cite{xu2025speculative} interleaves student
and teacher sampling. UNDO~\cite{jain2025undo} iteratively
prompts the teacher for rationales targeted at student
errors. \begingroup
Local Naturalness~\cite{just2025localnaturalness}
ranks responses using student likelihood at the segment level.
{GRAPE~\cite{zhang2025instructionfit} selects the response with
the highest normalized probability under the target model.
This instruction-tuning method is distinct from GRAPE for
robot-policy preference alignment~\cite{zhang2024grape}.
Rank-Surprisal Ratio~\cite{yang2026rsr} combines token rank
and surprisal to account for both alignment and informativeness.
BPR instead restricts scoring to the initial corrective steps
after the student's first divergence, rather than scoring
the full trajectory.} Our pipeline additionally uses behavioral
failure profiling to guide candidate generation.
\par
\endgroup

\paragraph{Error Aware Data Construction.}
A growing body of work recognizes that incorrect solutions
themselves contain valuable learning signals.
LeMa~\cite{an2023lema}, LEMMA~\cite{pan2025lemma}, and
Tong et al. \cite{tong2024can} pair incorrect reasoning with
teacher-generated corrections or error-type annotations.
TrajFusion~\cite{deng2026trajfusion} interleaves incorrect
trajectories with reflection prompts and correct
trajectories. Process supervision
approaches~\cite{lightman2024lets,wang2024mathshepherd,qi2025mutual,guan2025rstar,xi2024enhancing}
score intermediate steps or explore error nodes in
reasoning trees. Another
line~\cite{zhang2024score,han2024small,yan2025s3cmath}
collects erroneous trajectories by sampling at elevated
temperatures, which tends to produce nonsensical errors
unlikely to arise during normal
generation~\cite{lu2024step,renze-2024-effect}. Across
these methods, errors are treated uniformly regardless of
the failure pattern, and corrective trajectories follow
teacher patterns the student cannot reliably reproduce.
Our method addresses both through behavioral profiling
that classifies errors into distinct failure modes and
branching point reachability scoring that favors
corrections more likely to be learnable by the student.

\paragraph{Self-Correction in LLMs.}
One line of work uses external feedback from
verifiers or critic models for
self-correction~\cite{li2023reflection,tong2024can,shinn2023reflexion,li2024dotamath,du2024multiagent}.
GLoRe~\cite{havrilla2024glore} trains global and local
refinement models guided by stepwise outcome reward models.
Another line focuses on self-correction without external
feedback~\cite{zhang2024score,han2024small,yan2025s3cmath,wu2024large,lu2024step},
although Huang et al. \cite{huang2023large} demonstrate that LLMs
cannot reliably self-correct their reasoning without
external input. On the training side,
SCoRe~\cite{kumar2025score} proposes multi-turn online RL
for self-correction and shows that SFT on offline
correction traces fails due to distribution mismatch.
SuperCorrect~\cite{yang2025supercorrect} uses cross-model
collaborative DPO to optimize student self-correction with
teacher thought templates. Rather than training the model
to correct errors at inference time through RL or DPO, our
method constructs corrective training data filtered by
branching point reachability under the student, thereby
mitigating the distribution mismatch identified by SCoRe
while requiring only standard SFT.

\section{Method}\label{sec:method}

\begin{figure}[t]
\centering
\includegraphics[width=\linewidth]{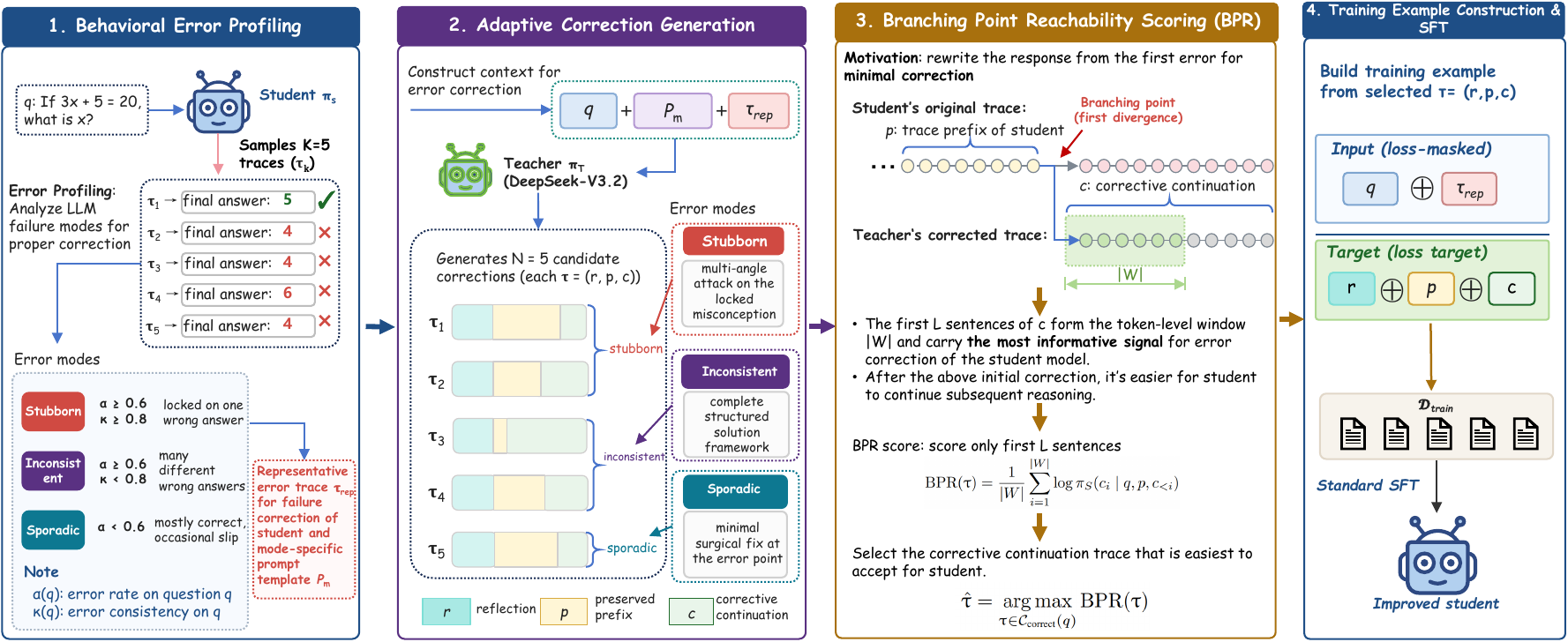}
\vspace{-1.2em}
\caption{Overview of our pipeline. Stage~1 profiles the student via repeated sampling and classifies each problem as stubborn, inconsistent, or sporadic from error rate $\alpha$ and consistency $\kappa$. Stage~2 prompts the teacher with a mode-specific template $P_m$ to generate $N$ candidates, each decomposed into reflection $r$, preserved prefix $p$, and corrective continuation $c$. The three mode-specific branches in Stage~2 illustrate alternative prompting strategies. For each problem, all $N$ candidates are generated using only the prompt corresponding to its diagnosed mode. The sampled-answer example in Stage~1 corresponds to the inconsistent mode. Stage~3 filters by correctness and selects the candidate maximizing branching point reachability over the first $L$ sentences of $c$.}
\label{fig:method}
\end{figure}

\subsection{Problem Formulation}\label{sec:formulation}

Let $\pi_T$ and $\pi_S$ denote the token-level conditional distributions of the teacher and student models, respectively, and let $q$ denote a mathematical problem. Standard training on teacher-generated traces optimizes the parameters $\theta$ of the student by minimizing
\begin{equation}\label{eq:sft}
\mathcal{L}_{\text{SFT}}(\theta) = -\mathbb{E}_{(q,\uptau) \sim \mathcal{D}_{\text{train}}} \left[\sum_{t=1}^{\ell} \log \pi_S(s_t \mid q, s_{<t})\right],
\end{equation}
where $\mathcal{D}_{\text{train}}$ is a dataset of correct reasoning traces generated by the teacher, each trace $\uptau = (s_1, \ldots, s_\ell)$ is a sequence of $\ell$ tokens, and $s_{<t} = (s_1, \ldots, s_{t-1})$ denotes all tokens preceding position $t$. However, reasoning traces exhibit strong path dependence: the student's ability to reproduce a step $s_t$ depends on the preceding context $(q, s_{<t})$ and on whether the student can reliably generate that context. When the reasoning style of the teacher differs from that of the student, per-step deviations accumulate over long chains, causing the student to encounter states at inference time that were never seen during training. This motivates a shift in the data construction objective: rather than seeking trajectories that are merely correct, we seek trajectories that are simultaneously correct and reachable by the student. Here, \emph{correct} means that the trace leads to the right answer, while \emph{reachable} means that the trace lies within the high-probability region of the distribution of the student, so that each step is a plausible continuation that the student could generate on its own. Let $\mathcal{C}_{\text{correct}}(q)$ denote the set of all correct reasoning traces for a given problem. The ideal selection objective is:
\begin{equation}\label{eq:ideal}
\hat\uptau = \argmax_{\uptau \in \mathcal{C}_{\text{correct}}(q)} \pi_S(\uptau \mid q).
\end{equation}
Our method approximates this objective through a generate-then-select pipeline illustrated in Figure~\ref{fig:method}, where the teacher proposes candidates and the distribution of the student itself serves as the selection criterion.

\subsection{Stage 1: Behavioral Error Profiling}\label{sec:profiling}

Not all errors are alike: different failure modes demand qualitatively different corrections. Given a dataset $\mathcal{D} = \{(q_i, y_i)\}_{i=1}^{M}$ of $M$ problems with ground-truth answers $y$, we draw $K$ reasoning traces $\uptau_1, \ldots, \uptau_K$ from $\pi_S$ for each problem $q$ and compute two diagnostic statistics.

\paragraph{Error rate.} The fraction of the $K$ samples that produce an incorrect final answer:
\begin{equation}
\alpha(q) = \frac{1}{K}\sum_{k=1}^{K} \mathbf{1}\left[\text{answer}(\uptau_k) \neq y\right],
\end{equation}
where $\text{answer}(\uptau_k)$ extracts the final answer from the $k$-th sampled trace.

\paragraph{Error consistency.} Among the incorrect samples, the fraction accounted for by the most frequent wrong answer $e$:
\begin{equation}
\kappa(q) = \max_{e \neq y} \frac{\sum_{k} \mathbf{1}[\text{answer}(\uptau_k) = e]}{\sum_{k} \mathbf{1}[\text{answer}(\uptau_k) \neq y]},
\end{equation}
where the maximization ranges over all incorrect answers $e$ that appear in the $K$ samples. A high $\kappa$ indicates that the errors of the student concentrate on a single wrong answer, while a low $\kappa$ indicates scattered, diverse failures.

Problems with $\alpha(q) = 0$ yield no error trajectory to diagnose and are excluded from the corrective pipeline, so the remainder of this section restricts attention to problems with $\alpha(q) > 0$. {We do not add a separate pool of problems for which all sampled responses are correct.} On this filtered set, $(\alpha, \kappa)$ partition errors into three distinct modes. Table~\ref{tab:error_modes} summarizes the classification criteria and the corresponding correction strategy.
This two-dimensional characterization moves beyond existing error taxonomies that classify errors by semantic type, such as calculation error or formula confusion. Our profiling instead captures the behavioral pattern of errors from the perspective of the student, which directly informs how corrections should be structured.

\begin{table}[t]
\centering
\caption{Error mode classification based on $(\alpha, \kappa)$ and the corresponding correction strategy.}
\label{tab:error_modes}
\footnotesize
\begin{tabular}{lccl}
\toprule
Mode & $\alpha$ & $\kappa$ & Correction Strategy \\
\midrule
Stubborn      & $\geq 0.6$ & $\geq 0.8$ & Multi-angle attack on the locked error path \\
Inconsistent  & $\geq 0.6$ & $< 0.8$  & Complete structured solution framework \\
Sporadic      & $0 < \alpha < 0.6$  & --         & Minimal surgical fix at the error point \\
\bottomrule
\end{tabular}
\end{table}

For each profiled problem, we additionally retain a single representative error trajectory $\uptau_{\text{rep}}$, which serves as the concrete instance of failure to be corrected by the teacher.
For stubborn errors, we select $\uptau_{\text{rep}}$ from the incorrect traces whose final answer matches the most frequent wrong answer. For inconsistent and sporadic errors, we randomly select an incorrect trace. After profiling, we obtain the error-annotated dataset:
\begin{equation}\label{eq:profiled}
\mathcal{D}_{\text{prof}} = \{(q, y, \{\uptau_k\}_{k=1}^{K}, m, \uptau_{\text{rep}}) \mid (q, y) \in \mathcal{D},\ \alpha(q) > 0\},
\end{equation}
where $m \in \{\texttt{stubborn}, \texttt{inconsistent}, \texttt{sporadic}\}$ denotes the error mode diagnosed using the criteria in Table~\ref{tab:error_modes}.

\subsection{Stage 2: Adaptive Correction Generation}\label{sec:generation}

\begingroup

Given the diagnosed error mode $m$, the teacher $\pi_T$ generates $N$ correction candidates using a mode-specific prompt $P_m$, the problem $q$, and the representative error trajectory $\uptau_{\text{rep}}$. For inconsistent errors, \texttt{student\_reasoning\_attempts} is a singleton list containing $\uptau_{\text{rep}}$. The templates are provided in Appendix~\ref{appendix:prompts}. Each prompt requests $N$ \emph{different} trajectories in one structured call, using mode-specific diversification instructions.
\par
\endgroup

The key design principles differ across modes. For stubborn errors, the prompt requests $N$ trajectories that attack the locked-in reasoning path from different angles, such as a direct fix, an alternative method, and a conceptual reframe. It explicitly asks the teacher to analyze the trap, conceptual pivot, fix, and recurrence-prevention check, and to distill this analysis into a concise first-person reflection. For inconsistent errors, the diverse incorrect answers suggest a lack of a stable problem-solving strategy, so the prompt requests $N$ structured solutions that supply different problem-solving frameworks, such as algebraic equations, step-by-step arithmetic, and work-backwards reasoning. For sporadic errors, the prompt instructs maximal preservation of the correct prefix of the student with minimal modification at the identified error point, and the $N$ trajectories vary only in their local correction style.

Each candidate consists of three parts: a reflection $r$, a preserved prefix $p$, and a corrective continuation $c$. The reflection $r$ is a concise first-person diagnosis that names where the reasoning of the student first went wrong. {The preserved prefix $p$ is the verbatim portion of $\uptau_{\text{rep}}$ up to but excluding the divergent step, and is empty when no correct prefix can be retained. With the singleton input used in our experiments, the prefix is the longest correct prefix of $\uptau_{\text{rep}}$, and the divergence point is its first erroneous step.} The corrective continuation $c$ extends $p$ to a correct final answer in a style natural to the student, or starts from scratch when $p$ is empty. Both the localization that yields $p$ and the diversification across the $N$ candidates are performed jointly by the teacher in a single structured call. Formally, the candidate set is:
\begin{equation}\label{eq:candidates}
\mathcal{C}(q) = \{(r_n, p_n, c_n)\}_{n=1}^{N} \sim \pi_T(\cdot \mid P_m, q, \uptau_{\text{rep}}).
\end{equation}

\subsection{Stage 3: Branching Point Reachability Scoring}\label{sec:selection}

Among the $N$ candidates, we select the one that the student model is most likely to reproduce.

\paragraph{Branching point reachability scoring.} We first discard candidates whose final answer does not match the ground truth $y$, and let $\mathcal{C}_{\text{correct}}(q) \subseteq \mathcal{C}(q)$ denote the resulting filtered set. For each remaining candidate $\uptau = (r, p, c) \in \mathcal{C}_{\text{correct}}(q)$, we extract a local window $W$ covering the first $L$ sentences of the corrective continuation $c$. BPR scores only $c$, since $r$ is a short reflection used as a training scaffold and $p$ is the output of the student. Let $|W|$ denote the number of tokens in this window, and let $c_i$ denote the $i$-th token of $c$, with $c_{<i} = (c_1, \ldots, c_{i-1})$ representing all preceding tokens within the continuation. The branching point reachability (BPR) score is the average log-probability of the window tokens under the student model, conditioned on the problem $q$ and the preserved prefix $p$:
\begin{equation}\label{eq:bpr}
\mathrm{BPR}(\uptau) = \frac{1}{|W|}\sum_{i=1}^{|W|} \log \pi_S(c_i \mid q, p, c_{<i}).
\end{equation}
This score restricts the log-probability computation to the local window immediately following the point where the reasoning of the student diverged. The rationale is that tokens later in the corrective continuation benefit from an already corrected context and receive systematically higher log-probabilities regardless of which candidate generated them. By focusing on the initial corrective steps, BPR isolates the signal that most directly reflects whether the student can execute the critical reasoning transition. The selected trajectory is:
\begin{equation}\label{eq:selection}
\hat\uptau = \argmax_{\uptau \in \mathcal{C}_{\text{correct}}(q)} \mathrm{BPR}(\uptau).
\end{equation}
In practice, we set $L\!=\!8$ sentences and extract sentence boundaries using newline delimiters, which are common in step by step mathematical reasoning, with a fallback to punctuation-based splitting. The BPR score is computed via a single forward pass through $\pi_S$ with no gradient computation.

\paragraph{Connection to the ideal objective.}
The ideal selection criterion in Eq.~\ref{eq:ideal}
maximizes $\pi_S(\uptau \mid q)$, or equivalently the
chain-rule decomposition
$\log \pi_S(\uptau \mid q) = \sum_{i=1}^{|\uptau|} \log \pi_S(\uptau_i \mid q, \uptau_{<i})$.
BPR is a deliberately biased proxy that departs from this
objective in two complementary ways. The restriction to a
local window $W$ rather than the full continuation is
justified above by the dilution argument and confirmed
empirically in Figure~\ref{fig:logprob_curve} of Appendix~\ref{appendix:logprob}. The other
divergence is that BPR averages rather than sums:
dividing by $|W|$ removes the length penalty that would
otherwise systematically favor shorter candidates regardless
of their per-token compatibility with the student. BPR
is therefore not a calibrated estimator of
$\log \pi_S(\uptau \mid q)$, but instead concentrates the
scoring signal on the segment that empirically governs
whether the student can execute the critical reasoning
transition, which is the property that ultimately matters
for selection.

\subsection{Training}\label{sec:training}

\begingroup

{Within each construction pass, the three stages select at most one training trajectory per retained question.} {The final SFT corpus uses two input-target formats. For 90\% of the examples, the input is $q \oplus \uptau_{\text{rep}}$ and the target is $r \oplus p \oplus c$. For the remaining 10\%, the input is $q$ alone and the target is the complete corrected solution $p \oplus c$, without the reflection or erroneous trace. These auxiliary examples come from the same $\alpha(q)>0$ problem pool.} {Algorithm~\ref{alg:pipeline} in Appendix~\ref{appendix:algorithm} summarizes the construction stages.}
\par
\endgroup

\clearpage
\section{Experiments}\label{sec:experiments}

\subsection{Experimental Setup}\label{sec:setup}

We use DeepSeek-V3.2~\citep{deepseekai2025deepseekv32pushingfrontieropen}
as teacher and Llama-3.1-8B-Instruct~\citep{dubey2024llama}
and DeepSeek-Math-7B-Instruct~\citep{shao2024deepseekmath}
as students. {Each construction pass profiles all {394.99K}
MetaMathQA instances~\citep{yu2024metamath} with $K=5$
student samples and generates $N=5$ teacher candidates
only for instances with {$\alpha(q)>0$}, as detailed in Appendix~\ref{appendix:algorithm}.
The complete first pass yields 62.74K and 65.76K examples
for Llama and DeepSeek. A fresh full second pass is run.
Part of its output is merged, prioritizing previously unseen
questions until reaching 88.90K.}
SFT, RFT, S$^3$C-Math, and Ours use MetaMathQA;
GPTAug, RefAug, ISC, and LEMMA construct supervision from
GSM8K and MATH source problems using their respective recipes.
MetaMathQA is also derived from GSM8K and MATH. Sample counts
and filtering follow each method's
construction, as detailed in Appendix~\ref{appendix:baselines}.
Training settings and evaluation benchmarks are detailed in Appendix~\ref{appendix:benchmarks}.

% \begingroup
% \paragraph{Models and data.}
% We use DeepSeek-V3.2~\citep{deepseekai2025deepseekv32pushingfrontieropen}
% as teacher and Llama-3.1-8B-Instruct~\citep{dubey2024llama}
% and DeepSeek-Math-7B-Instruct~\citep{shao2024deepseekmath}
% as students. {Each construction pass profiles all {394.99K}
% MetaMathQA instances~\citep{yu2024metamath} with $K=5$
% student samples and generates $N=5$ teacher candidates
% only for instances with {$\alpha(q)>0$}, as detailed in Appendix~\ref{appendix:algorithm}.
% The complete first pass yields 62.74K and 65.76K examples
% for Llama and DeepSeek. A fresh full second pass is run.
% Part of its output is merged, prioritizing previously unseen
% questions until reaching 88.90K.}

% \paragraph{Comparison data.}
% SFT, RFT, S$^3$C-Math, and Ours use MetaMathQA;
% GPTAug, RefAug, ISC, and LEMMA construct supervision from
% GSM8K and MATH source problems using their respective recipes.
% MetaMathQA is also derived from GSM8K and MATH. Sample counts
% and filtering follow each method's
% construction, as detailed in Appendix~\ref{appendix:baselines}.

% \paragraph{Training and evaluation.}
% Training settings and evaluation benchmarks are detailed in Appendix~\ref{appendix:benchmarks}.
% \par\endgroup

\subsection{Main Results}\label{sec:main_results}

\begin{table*}[t]
\centering
\caption{Accuracy (\%) on in-distribution and out-of-distribution
benchmarks. Best results are in \textbf{bold} and second-best are \underline{underlined}. {The 88.90K Ours rows report mean $\pm$ sample standard deviation across three SFT seeds. Avg statistics are computed across per-seed five-benchmark averages.}}
\label{tab:main_results}
\small
\setlength{\tabcolsep}{2pt}
\resizebox{\linewidth}{!}{\begin{tabular}{l l c cc ccc c}
\toprule
& & & \multicolumn{2}{c}{\textbf{In-Distribution}}
& \multicolumn{3}{c}{\textbf{Out-of-Distribution}}
& \\
\cmidrule(lr){4-5} \cmidrule(lr){6-8}
\textbf{Student Model} & \textbf{Method} & \textbf{\# Samples}
& GSM8K & MATH & SVAMP & ASDiv & MAWPS & \textbf{Avg.} \\
\midrule
\multirow{9}{*}{Llama-3.1-8B}
& SFT       & 394.99K& 80.3 & 39.7 & 77.5 & 82.4 & 89.2 & 73.8 \\
& RFT~\citep{yuan2023scaling}       & 86.52K & 76.4 & 40.6 & 78.6 & 78.8 & 83.2 & 71.5 \\
& ISC~\citep{han2024small}       & 86.78K & 72.5 & 43.5 & 80.4 & 81.7 & 82.7 & 72.2 \\
& S$^3$C-Math~\citep{yan2025s3cmath} & 927K & 83.1 & 42.8 & 82.4 & --   & --   & --   \\
& GPTAug    & 88.62K & 73.3 & 41.6 & 81.0 & 81.8 & 86.3 & 72.8 \\
& RefAug~\citep{zhang-etal-2024-learn}    & 29.94K & 77.3 & 42.3 & 81.7 & 82.4 & 88.7 & 74.5 \\
& LEMMA~\citep{pan2025lemma}     & 88.90K & 80.4 & \textbf{48.8} & 83.7 & 84.5 & 88.9 & 77.3 \\
\cmidrule(lr){2-9}
& FPRT (Ours)      & 62.74K & \underline{85.5} & 44.3 & \underline{86.4} & \underline{84.9} & \textbf{93.9} & \underline{79.0} \\
& FPRT (Ours) & {88.90K} & {\textbf{86.2$\pm$0.4}} & {\underline{46.2$\pm$0.5}} & {\textbf{86.8$\pm$0.2}} & {\textbf{86.1$\pm$0.3}} & {\underline{93.3$\pm$0.4}} & {\textbf{79.7$\pm$0.4}} \\
\midrule
\multirow{9}{*}{DeepSeek-Math-7B}
& SFT       & 394.99K& 79.4 & 42.0 & 79.4 & 87.8 & 90.2 & 75.8 \\
& RFT       & 86.52K & 77.5 & 43.9 & 77.4 & 86.9 & 89.6 & 75.1 \\
& ISC       & 86.78K & 72.2 & 39.5 & 75.6 & 84.4 & 89.8 & 72.3 \\
& S$^3$C-Math & 927K & 82.8 & 42.6 & 83.1 & --   & --   & --   \\
& GPTAug    & 88.62K & 78.6 & 46.0 & 74.2 & 88.9 & 89.9 & 75.5 \\
& RefAug    & 29.94K & 77.0 & 39.8 & 73.7 & 83.8 & 85.3 & 71.9 \\
& LEMMA     & 88.90K & 81.3 & \textbf{50.7} & 81.7 & 89.4 & 91.1 & 78.8 \\
\cmidrule(lr){2-9}
& FPRT (Ours)      & 65.76K & \underline{83.6} & 47.1 & \underline{83.9} & \textbf{90.1} & \textbf{95.3} & \underline{80.0} \\
& FPRT (Ours) & {88.90K} & {\textbf{84.0$\pm$0.2}} & {\underline{48.9$\pm$0.3}} & {\textbf{85.4$\pm$0.2}} & {\underline{89.8$\pm$0.4}} & {\underline{94.7$\pm$0.3}} & {\textbf{80.6$\pm$0.3}} \\
\bottomrule
\end{tabular}}
\end{table*}

\begingroup
\paragraph{Consistent Average Lead across Two Student Backbones.}
Table~\ref{tab:main_results} reports accuracy on the two student models. Our method attains the highest average on both backbones, 79.7\% on Llama-3.1-8B and 80.6\% on DeepSeek-Math-7B, exceeding the strongest comparison method, LEMMA, by 2.4\% and 1.8\%, respectively. This lead persists at the smaller 62.74K and 65.76K configurations, where the averages reach 79.0\% and 80.0\% and still surpass every comparison method with a reported five-benchmark average. The ranking is therefore stable across the two corpus sizes rather than tied to a single training budget. The improvement also holds across a general purpose and a math specialized backbone, indicating that the gain is not tied to any single pretraining distribution. The average-accuracy margins over RFT and ISC range from 5.5\% to 8.3\%.

\paragraph{Larger Margin under Distribution Shift.}
The improvement is concentrated on the three out-of-distribution benchmarks, where the averages of 88.7\% on Llama-3.1-8B and 90.0\% on DeepSeek-Math-7B exceed LEMMA by 3.0\% and 2.6\%, respectively. On DeepSeek-Math-7B, the in-distribution average rises only modestly from 66.0\% for LEMMA to 66.5\% for our method, indicating that most of the 1.8\% overall lead originates from out-of-distribution gains. MAWPS in particular shows leads over LEMMA of 4.4\% and 3.6\% on the two student models. This pattern is consistent with the hypothesis that anchoring corrections at the first divergence yields trajectories that transfer beyond MetaMathQA style templates.

\paragraph{Trade Off on MATH.}
On MATH, our method trails LEMMA by 2.6\% on Llama-3.1-8B and 1.8\% on DeepSeek-Math-7B, while ranking second on both backbones. One possible explanation is that LEMMA's error-type grounded mistake augmentation provides supervision better suited to MATH. Correctness filtering may also limit the available teacher candidates on difficult problems. These hypotheses require further analysis of error-type coverage and candidate retention rates; the current results do not isolate candidate construction from candidate selection as the source of the gap.

\paragraph{Favorable Sample Efficiency.}
{We evaluate our method on two different training corpus sizes with the same pipeline. The 62.74K and 65.76K configurations use the complete first-pass corpora and are subsets of the corresponding 88.90K corpora assembled from two passes. They are used for the ablation and window size studies in Section~\ref{sec:ablation}.}
At these smaller subsets, roughly 70\% of the LEMMA corpus and about one sixth and one fifteenth of the SFT and S$^3$C-Math corpora, our method already exceeds every comparison method with a reported five-benchmark average, surpassing LEMMA by 1.7\% and 1.2\%. Scaling to 88.90K, comparable to the 88.90K of LEMMA, lifts the margin to 2.4\% and 1.8\%, so the lead does not collapse once both methods operate at the same data budget.

\par\endgroup

\subsection{Ablation Studies}\label{sec:ablation}

\begingroup
\paragraph{Supervision controls.}
Table~\ref{tab:supervision_controls} compares two controls
on the same first-pass problem sets, with teacher, student,
example count, checker, and SFT settings matched to Ours.
Direct teacher-trace SFT uses one verified-correct
DeepSeek-V3.2 solution per question, generated from the
question alone and resampled if incorrect. Our pipeline
outperforms this control by 3.1\% on Llama and 1.9\% on
DeepSeek, evaluating the full supervision-construction pipeline.

Random-one-correct follows the same profiling, candidate pools,
and correctness filtering as Ours, but uniformly selects
one correct candidate per retained problem without scoring.
It achieves 76.6\% and 79.0\%, compared with BPR's
79.0\% and 80.0\% on Llama and DeepSeek, respectively.
\par\endgroup

\begin{table}[t]

\centering
\caption{Accuracy (\%) of supervision controls on the same first-pass problem sets, with one correct training trajectory per problem. Random-one-correct and BPR adopt the same correct candidate pools. Direct teacher-trace SFT generates solutions from the question alone. Best accuracy within each student is bold.}
\label{tab:supervision_controls}
\small
\setlength{\tabcolsep}{4pt}
\resizebox{\linewidth}{!}{\begin{tabular}{l l cc ccc c}
\toprule
\textbf{Student Model} & \textbf{Method}
& GSM8K & MATH & SVAMP & ASDiv & MAWPS & \textbf{Avg.} \\
\midrule
\multirow{3}{*}{Llama-3.1-8B}
& Direct teacher-trace SFT & 84.0 & 42.7 & 82.0 & 82.1 & 88.5 & 75.9 \\
& {Random-one-correct} & {81.6} & {42.1} & {84.7} & {81.2} & {93.2} & {76.6} \\
\cmidrule(lr){2-8}
& BPR selection (Ours) & \textbf{85.5} & \textbf{44.3} & \textbf{86.4} & \textbf{84.9} & \textbf{93.9} & \textbf{79.0} \\
\midrule
\multirow{3}{*}{DeepSeek-Math-7B}
& Direct teacher-trace SFT & 82.6 & 46.5 & 81.5 & 89.2 & 90.7 & 78.1 \\
& {Random-one-correct} & {83.0} & {46.8} & {82.6} & {89.6} & {92.8} & {79.0} \\
\cmidrule(lr){2-8}
& BPR selection (Ours) & \textbf{83.6} & \textbf{47.1} & \textbf{83.9} & \textbf{90.1} & \textbf{95.3} & \textbf{80.0} \\
\bottomrule
\end{tabular}}
\end{table}

\begingroup
\paragraph{Profiling Drives the Bulk of the Gain, with BPR Refining on Top.}
To isolate the contributions of behavioral profiling and branching point reachability scoring, we train three reduced variants of our pipeline against the full method, each removing one or both components while holding the rest of the training procedure fixed. These ablations use the Llama-3.1-8B first-pass configuration. The first replaces behavioral profiling with a uniform minimal-correction prompt template applied to all errors, preserving the longest correct prefix and changing only the erroneous reasoning, the second replaces BPR with global log-probability scoring over the full correction, and the third combines both ablations. Results across the five benchmarks are summarized in Table~\ref{tab:component_ablation}. Replacing profiling with a uniform prompt causes the largest single drop, from 79.0\% to 72.9\% on average, with the largest benchmark-specific drop on GSM8K (8.0\%). Replacing BPR with global log-probability scoring reduces the average from 79.0\% to 77.3\%, a smaller decline in average accuracy that matches the dilution argument in Section~\ref{sec:selection} and Figure~\ref{fig:logprob_curve}. MAWPS exhibits a slight reversal of 0.3\% in favor of global scoring, 94.2\% against 93.9\%. Removing both components yields 72.4\%, only 0.5\% below the variant that keeps BPR alone. BPR adds 1.7\% when profiling supplies a tailored candidate pool, suggesting that its benefit depends on candidate construction. These results suggest complementary contributions from the two components: profiling determines the shape of the corrective trajectory, while BPR ranks the candidates within that pool by reachability under the student.
{The exact uniform prompt is in Appendix~\ref{appendix:prompt_uniform}.}
\par\endgroup

\begin{table*}[t]
\centering
\caption{Accuracy (\%) of component ablations across five benchmarks: GSM8K, MATH, SVAMP, ASDiv, and MAWPS. ``Uniform
prompting'' replaces behavioral profiling with a single
prompt template for all errors. ``Global scoring'' replaces
BPR with global log-probability scoring over the full
correction.}
\label{tab:component_ablation}
\small
\begin{tabular*}{\linewidth}{@{\extracolsep{\fill}}l c c c c c c}
\toprule
\textbf{Variant} & GSM8K & MATH & SVAMP & ASDiv & MAWPS & \textbf{Avg.} \\
\midrule
Full method                              & \textbf{85.5} & \textbf{44.3} & \textbf{86.4} & \textbf{84.9} & 93.9 & \textbf{79.0}  \\
\quad w/o profiling (uniform prompting)  & 77.5 & 38.3 & 80.4 & 78.7 & 89.4 & 72.9 \\
\quad w/o BPR (global scoring)           & 82.3 & 42.5 & 85.6 & 81.9 & \textbf{94.2} & 77.3 \\
\quad w/o both (uniform + global)        & 75.3 & 37.6 & 79.8 & 79.6 & 89.8 & 72.4 \\
\bottomrule
\end{tabular*}
\end{table*}

\begin{samepage}
\begingroup
\paragraph{Boundary localization.}
{An audit of 50 traces from the 65.76K DeepSeek configuration
finds 88\% accurate first-error boundaries.
Perturbing BPR boundaries on 12\% of questions with preserved
prefixes changes average accuracy from 80.0\% to 79.7\%
in a single run. Appendix~\ref{appendix:localization_results}
gives the audit and perturbation protocol.}
\par\endgroup
\end{samepage}

\begin{figure}[t]
    \vspace{-1em}
    \centering
    \begin{minipage}[t]{0.48\columnwidth}
        \centering
        \includegraphics[width=0.92\linewidth]{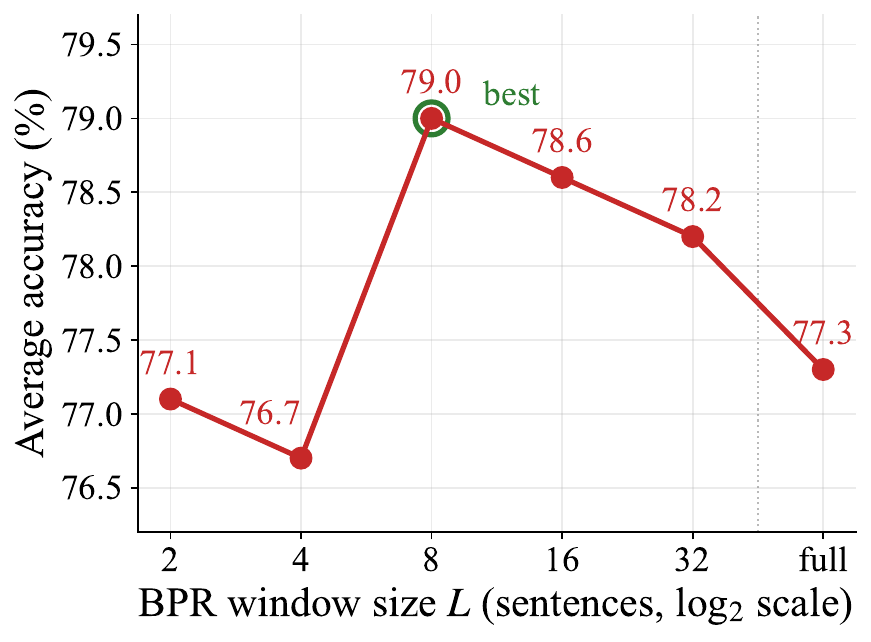}
        \caption{The effect of BPR window size $L$ on average accuracy.}
        \label{fig:window_size_avg}
    \end{minipage}
    \hfill
    \begin{minipage}[t]{0.48\columnwidth}
        \centering
        \includegraphics[width=0.92\linewidth]{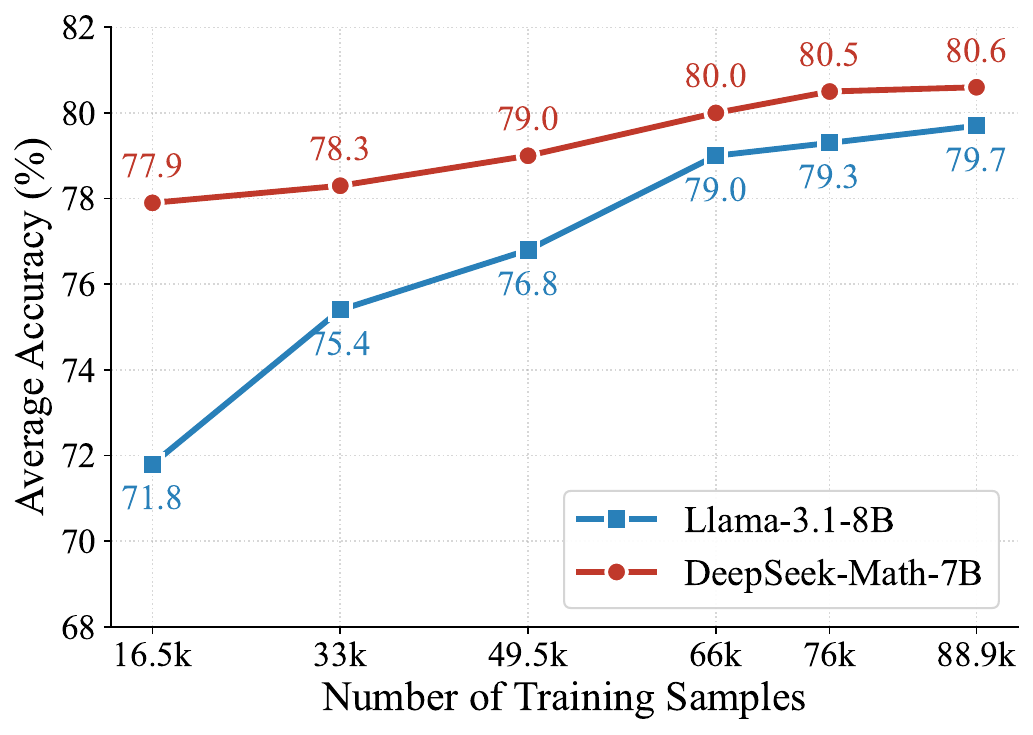}
        \caption{Data scaling: average accuracy across five benchmarks.}
        \label{fig:data_scaling_avg}
    \end{minipage}
    \par\vspace{-1.3em}
\end{figure}

\begingroup
\paragraph{Robust to Window Size with a Moderate Optimum.}
To probe the sensitivity of BPR to its only hyperparameter, we sweep the window size $L$ across $\{2, 4, 8, 16, 32, \text{full}\}$ sentences while keeping the rest of the pipeline fixed, where the full setting recovers global scoring over the entire correction. The resulting average accuracy curve is plotted in Figure~\ref{fig:window_size_avg}. Average accuracy peaks at 79.0\% when $L=8$, and we adopt this value as the default in all other experiments. The curve is markedly asymmetric around the optimum. For $L<8$, accuracy is 77.1\% and 76.7\%, indicating that very short windows do not contain enough tokens to discriminate reachable from unreachable trajectories. For $L>8$, accuracy decays monotonically, falling to 78.6\% at $L=16$, 78.2\% at $L=32$, and 77.3\% at the global setting. This 1.7\% drop matches the gap between BPR and global scoring reported in the component ablation. The decay to the right of the optimum is consistent with the dilution argument in Section~\ref{sec:selection} and Figure~\ref{fig:logprob_curve}, since enlarging the window past the divergence transition admits tokens whose high probability, inherited from an already corrected context, obscures the discriminative signal at the branching point. The shape of the curve, with both very short and global windows lying below $L=8$, identifies a moderate window as the regime where BPR captures the intended signal. This regime is broad enough to smooth local noise yet narrow enough to isolate the critical transition. A complementary token-level analysis of where the discriminative signal concentrates is provided in Figure~\ref{fig:logprob_curve} of Appendix~\ref{appendix:logprob}.

\paragraph{Data Scaling.}
To examine how the size of our constructed corpus translates into student accuracy, we scale the training set from 16.5K to 88.90K examples for both Llama-3.1-8B and DeepSeek-Math-7B and retrain each student under the same recipe. Intermediate sizes uniformly subsample the 88.90K corpus. Each configuration uses one SFT run with the same fixed seed; the 88.90K Ours rows in Table~\ref{tab:main_results} instead report three-seed means. The resulting accuracy curves are plotted in Figure~\ref{fig:data_scaling_avg}. Both student models improve monotonically with more data, yet their slopes differ in a manner that reflects their starting competence. Llama-3.1-8B climbs from 71.8\% to 79.7\%, a total gain of 7.9\%. The gains at the final two plotted sizes are 0.3\% and 0.4\%, respectively. DeepSeek-Math-7B starts at 77.9\% and reaches 80.6\%, a smaller gain of 2.7\%. The final two increments are 0.5\% and 0.1\%, respectively. {The contrast suggests that the general purpose backbone benefits substantially from targeted corrective trajectories, which supply mathematical patterns under-represented in its pretraining. The math specialized backbone, by contrast, already encodes much of the relevant signal and extracts only incremental value from additional supervision. The flattening at the upper end of both curves indicates that the marginal utility of scaling our corpus diminishes once the dominant failure modes in the training distribution have been absorbed.} Within the evaluated range, both curves show smaller marginal gains at the larger training sizes.
\par\endgroup

\begingroup
\paragraph{{Seeds, teachers, and inference behavior.}}
{Table~\ref{tab:main_results} reports three-seed results for each fixed 88.90K corpus (the original SFT seed and two additional seeds), with all other settings fixed.}
{In a matched single-run comparison using the same fixed SFT seed, changing only the teacher to DeepSeek-V4-Flash gives 81.0\% on DeepSeek, versus 80.6\% with DeepSeek-V3.2.}
Full results and protocols are in
Appendices~\ref{appendix:robustness_results}
and~\ref{appendix:inference_results}.
\par\endgroup

\section{Conclusion}\label{sec:conclusion}

We present a reasoning-distillation pipeline that combines behavioral failure profiling, mode-specific correction generation, and branching point reachability scoring. Profiling tailors supervision to how the student fails, while localized scoring selects correct candidates whose initial corrective steps are more reproducible by the student. Across GSM8K, MATH, and three out-of-distribution benchmarks, the method achieves the highest average accuracy on both Llama-3.1-8B and DeepSeek-Math-7B. Ablations support the contributions of profiling and localized scoring. The results highlight the value of jointly considering student failure patterns and corrective-trajectory reachability when constructing distillation data.

\begingroup
\section{Limitations}\label{sec:limitations}

\vspace{-0.8em}
\paragraph{Dependence on a strong teacher model.}
Our pipeline relies on a capable teacher $\pi_T$, instantiated
as DeepSeek-V3.2 in the main experiments, both to generate correction candidates and to
localize the divergent step in the error trajectory of the
student. The quality and diversity of the constructed training
data are therefore upper-bounded by the competence of the
teacher on the target problem distribution. When the
teacher itself fails on a problem, no usable candidate enters
$\mathcal{C}_{\text{correct}}(q)$ and the problem is dropped
from the final training set.

\vspace{-0.8em}
\paragraph{Sensitivity to the sampling budget $K$.}
Behavioral profiling estimates $(\alpha, \kappa)$ from $K\!=\!5$
student samples per problem. With small $K$, both statistics
are coarse-grained: $\alpha$ takes values in
$\{0, 0.2, 0.4, 0.6, 0.8, 1.0\}$ and $\kappa$ takes a similarly
small discrete set. This grants robustness to threshold choice,
as shown in Appendix~\ref{appendix:threshold}, but also limits the resolution of
the diagnosed mode. Larger $K$ would yield finer profiling at
proportionally higher sampling cost, but we did not explore
this trade-off systematically.
\par\endgroup

\section*{Acknowledgments}
This work was supported by National Natural Science Foundation of China (No.62302523) and Natural Science Foundation of Hunan Province (No.2026JJ50231). This work was carried out in part using computing resources at the HPC Center of Central South University.

% \clearpage
{
\small
\bibliographystyle{plain}
% \bibliography{custom}

}

\newpage
\appendix

\section{Full Pipeline Algorithm}\label{appendix:algorithm}

\begingroup
{Algorithm~\ref{alg:pipeline} shows one construction pass followed by SFT. For the two-pass setting, both passes use the same original student checkpoint, and SFT is performed after the final corpus is assembled.}

\begin{algorithm}[h]

\caption{{Full Pipeline}}
\label{alg:pipeline}
\begin{algorithmic}[1]
\Require Dataset $\mathcal{D} = \{(q_i, y_i)\}_{i=1}^{M}$, teacher $\pi_T$, student $\pi_S$, hyperparameters $K$, $N$, $L$
\Ensure Fine-tuned student $\hat{\pi}_S$

\State $\mathcal{D}_{\text{selected}} \leftarrow \emptyset$
\For{each $(q,y)\in\mathcal{D}$}
    \State Sample $K$ traces $\{\uptau_k\}_{k=1}^{K}\sim\pi_S(\cdot\mid q)$

    \State Compute $\alpha(q)$

    \If{$\alpha(q)=0$}
        \State \textbf{continue}
    \EndIf

    \State Compute $\kappa(q)$
    \State Classify error mode $m$ by Table~\ref{tab:error_modes}
    \State Select representative error trace $\uptau_{\text{rep}}$
    \State {Generate $N$ candidate trajectories in one teacher call: $\{(r_n,p_n,c_n)\}_{n=1}^{N}\sim\pi_T(\cdot\mid P_m,q,\uptau_{\text{rep}})$}
    \State $\mathcal{C}_{\text{correct}}\leftarrow\{(r_n,p_n,c_n):\operatorname{answer}(c_n)=y\}$
    \If{$\mathcal{C}_{\text{correct}}=\emptyset$}
        \State \textbf{continue}
    \EndIf
    \For{each $\uptau=(r,p,c)\in\mathcal{C}_{\text{correct}}$}
        \State $W\leftarrow$ first $L$ sentences of $c$
        \State $\mathrm{BPR}(\uptau)\leftarrow \dfrac{1}{|W|}\displaystyle\sum_{i=1}^{|W|}\log\pi_S(c_i\mid q,p,c_{<i})$
    \EndFor
    \State $\hat\uptau \leftarrow \argmax_{\uptau \in \mathcal{C}_{\text{correct}}} \mathrm{BPR}(\uptau)$, with components $(\hat r, \hat p, \hat c)$

    \State Append $(q,\uptau_{\text{rep}},\hat r,\hat p,\hat c)$ to $\mathcal{D}_{\text{selected}}$
\EndFor

\State {Construct $\mathcal{D}_{\text{train}}$: 90\% use $(q\oplus\uptau_{\text{rep}},\,\hat r\oplus\hat p\oplus\hat c)$; 10\% use $(q,\,\hat p\oplus\hat c)$}
\State Fine-tune $\pi_S$ on $\mathcal{D}_{\text{train}}$ with loss on target tokens only
\State \Return $\hat{\pi}_S$
\end{algorithmic}
\end{algorithm}

\par\endgroup

\section{Training and Evaluation Details}\label{appendix:benchmarks}

\begingroup
\paragraph{Training formats.}\label{appendix:training}
{For 90\% of the final examples, the input is
$q\oplus\uptau_{\text{rep}}$ and the target is
$\hat r\oplus\hat p\oplus\hat c$. For the remaining 10\%,
the input is $q$ alone and the target is the complete
corrected solution $\hat p\oplus\hat c$, without reflection.
These auxiliary examples come from the same
$\alpha(q)>0$ pool, not a separate $\alpha(q)=0$ pool.
The input is loss-masked in both formats.}

{
\paragraph{Training settings.}
We use full-parameter SFT in LLaMA-Factory~\citep{zheng2024llamafactory}
for 2--3 epochs with {a learning rate between $1\times10^{-5}$ and $2\times10^{-5}$}, a cosine
schedule with 3\% warmup, an effective batch size of 128, and a maximum
sequence length of 2048.
Each SFT run uses one NVIDIA H200 or eight NVIDIA GeForce RTX 3090 GPUs.
\par}

{\paragraph{Evaluation.} We evaluate on GSM8K~\citep{cobbe2021training} and MATH~\citep{hendrycks2021measuring} for in-distribution, and SVAMP~\citep{patel2021nlp}, ASDiv~\citep{miao-etal-2020-diverse}, and MAWPS~\citep{koncel2016mawps} for out-of-distribution, using greedy decoding with exact-match accuracy after answer extraction.}

\par\endgroup

\clearpage
\section{Comparison Method Details}\label{appendix:baselines}

\begingroup
{We compare against SFT, RFT~\citep{yuan2023scaling},
ISC~\citep{han2024small}, S$^3$C-Math~\citep{yan2025s3cmath},
GPTAug, RefAug~\citep{zhang-etal-2024-learn}, and
LEMMA~\citep{pan2025lemma}. All baselines draw from the same
training data source for fair comparison: the GSM8K and MATH
training problems. They do not necessarily
use identical surface questions or constructed instances.}

\begingroup
\paragraph{{Methods applied to MetaMathQA.}}
{SFT uses the complete 394.99K MetaMathQA corpus of correct
solutions.} RFT rejection-samples the student and retains
correct self-generated traces, with at most one per
distinct question. {S$^3$C-Math injects intermediate mistakes
via step-level sampling and constructs reflective
corrections, adding approximately 532K instances to
MetaMathQA for approximately 927K samples in total.}
Our method profiles the student on MetaMathQA and
constructs corrections as in Appendix~\ref{appendix:algorithm}.
\par\endgroup

\begingroup
\paragraph{{Methods constructing supervision from source problems.}}
GPTAug generates direct teacher solutions, ISC constructs
self-correction sequences, RefAug enriches solutions with
reflection phases, and LEMMA performs error-type grounded
mistake augmentation. These methods follow their respective
construction recipes based on the GSM8K and MATH source
problems rather than using our MetaMathQA variants.
{The original RefAug configuration contains 29.94K samples;
LEMMA retains 88.90K trajectories after correctness
filtering.
The other sample counts follow the methods' original
construction settings.}
\par\endgroup

\begingroup
\paragraph{Controls on the same first-pass problems.}
The direct teacher-trace control uses exactly the problems
in our 62.74K Llama and 65.76K DeepSeek first-pass sets.
DeepSeek-V3.2 generates a solution conditioned only on the
question; its final answer is checked with the same checker
as our pipeline, and incorrect generations are resampled
until a correct trace is obtained. Each problem contributes
one verified-correct trajectory. Teacher, student, and SFT
settings match our method. This compares complete
supervision-construction pipelines rather than isolating BPR.
\par\endgroup

\par\endgroup

\section{Prompt Templates for Adaptive Correction Generation}\label{appendix:prompts}

This appendix provides the full prompt templates used by the
teacher model $\pi_T$ for each diagnosed error mode in
Stage~2 of our pipeline, described in Section~\ref{sec:generation}. All
three templates share the same JSON output schema, a list
of $N$ corrective trajectories with four fields:
\texttt{error\_step} for a first-person diagnosis,
\texttt{preserved\_prefix} for the verbatim correct portion of
the reasoning of the student that may be empty,
\texttt{correction} for the continuation in the style of the
student ending with ``The answer is: [value]'', and
\texttt{final\_answer} for the final numerical or symbolic answer.
\begingroup
The templates differ in their input
slots and in the strategies used to diversify the $N$
candidates, as detailed below.
The slot \texttt{\{trajectory\_examples\}} contains a fixed
block of six examples demonstrating the reflection $r$,
preserved prefix $p$, and corrective continuation $c$.
The same block is used across all three templates.
\par\endgroup

The slot \texttt{\{task\_data\}} fills the per-problem inputs into $P_m$:
the problem, correct answer, profiling statistics, answer-level summary,
and one representative incorrect trace. Stubborn and sporadic prompts use
\texttt{student\_reasoning}; the inconsistent prompt uses a singleton
\texttt{student\_reasoning\_attempts} list.

\clearpage
\par
\subsection{Prompt for Stubborn Errors}\label{appendix:prompt_stubborn}

For stubborn errors, the majority of the $K$ sampled
attempts of the student produce the same wrong answer,
indicating a deep-rooted conceptual flaw rather than a
random slip. The prompt therefore instructs the teacher to
attack the locked error path from multiple angles, while
still preserving any correct portion of the reasoning of
the student that precedes the divergence.

\begin{tcolorbox}[breakable, lines before break=4, boxsep=2pt, left=5pt, right=5pt, top=2pt, bottom=2pt, before skip=4pt, after skip=4pt, colback=gray!5, colframe=black!60, title={\textbf{Prompt Template for Stubborn Errors}}, fonttitle=\small]
\begingroup\small
\setlength{\parskip}{1pt}
You are an expert Math Corrector. You specialize in diagnosing SYSTEMATIC MISCONCEPTIONS.
\par
\textbf{The incorrect outcomes concentrate on one wrong answer.} \textbf{Attack the locked-in reasoning path from different angles:} direct correction, an alternative method, conceptual reframing, work backwards, or substitution-based verification. \textbf{Diagnose the trap}, the conceptual pivot, the fix, and a check that prevents recurrence. Distill that analysis into a concise first-person \texttt{error\_step}.
\par
\textbf{Correction-angle examples:} trajectory~1 applies a direct fix to the specific algebraic/logical error; trajectory~2 solves the problem via an alternative method entirely; trajectory~3 reframes the underlying concept before solving; further trajectories explore additional creative variations such as work-backwards or substitution-based verification.

\par\noindent \textbf{Generate exactly} \texttt{\{n\_candidates\}} \textbf{DIFFERENT corrective trajectories in ONE JSON array.} Each object must contain these string fields:
\begin{enumerate}[leftmargin=*, itemsep=0.5pt, parsep=0pt, topsep=1pt, partopsep=0pt]
\item \textbf{error\_step:} a concise first-person diagnosis naming where and why the first error occurred. For example, "I assumed that\ldots" or "I forgot to multiply by\ldots".
\item \textbf{preserved\_prefix:} the student's LONGEST correct prefix, copied VERBATIM from the beginning up to but excluding the first erroneous step; empty if none.
\item \textbf{correction:} continue from \texttt{preserved\_prefix} to a complete correct solution. Match the student's vocabulary, formatting, and tone. End with exactly "The answer is: [value]".
\item \textbf{final\_answer:} only the final numerical or symbolic answer.
\end{enumerate}
\par
\textbf{Localize the first error} and diversify the \texttt{\{n\_candidates\}} continuations jointly in this single call. \textbf{Never invent an error in correct reasoning.} Avoid apologies, meta-commentary such as "Let me reconsider" or "I made an error", and explanations outside the array.
\par
\textbf{Reference output-format examples:} \texttt{\{trajectory\_examples\}}

\textbf{Task data:} \texttt{\{task\_data\}}
\par\endgroup
\end{tcolorbox}
\par

\par
\subsection{Prompt for Inconsistent Errors}\label{appendix:prompt_inconsistent}

For inconsistent errors, the prompt emphasizes \emph{structured reasoning}
to provide a stable problem-solving framework. The slot
\texttt{student\_reasoning\_attempts} supports one or multiple trajectories.
With multiple trajectories, the teacher identifies their shared correct
prefix and where their reasoning diverges; with one, it preserves that
trajectory's longest correct prefix and identifies its first erroneous step.
In our experiments, the slot is a singleton list containing only
$\uptau_{\text{rep}}$. The answer-level summary covers all $K$
samples, without supplying their full reasoning chains.

\begin{tcolorbox}[breakable, lines before break=4, boxsep=2pt, left=5pt, right=5pt, top=2pt, bottom=2pt, before skip=4pt, after skip=4pt, colback=gray!5, colframe=black!60, title={\textbf{Prompt Template for Inconsistent Errors}}, fonttitle=\small]
\begingroup\small
\setlength{\parskip}{1pt}
You are an expert Math Corrector. You specialize in STABILIZING chaotic reasoning patterns.
\par
\textbf{The sampled outcomes are unstable.} \textbf{Supply a structured framework:} state the givens and goal, define variables and equations, calculate step by step, and verify the result. \textbf{Vary frameworks:} equations, verifiable arithmetic, working backwards, dimensional analysis, or unit tracking. The answer summary describes all sampled outcomes; \texttt{student\_reasoning\_attempts} may contain one or more trajectories to be analyzed. With multiple trajectories, preserve any shared correct prefix and identify where their reasoning diverges; with one, preserve that trajectory's longest correct prefix and diagnose its first erroneous step. If no correct prefix exists, leave \texttt{preserved\_prefix} empty. Do not infer a shared prefix from unprovided trajectories.
\par
\textbf{Framework examples:} trajectory~1 sets up and solves algebraic equations; trajectory~2 breaks the problem into small verifiable arithmetic sub-calculations; trajectory~3 works backwards from the answer structure; further trajectories use dimensional analysis, unit tracking, or diagram-based reasoning.

\par\noindent \textbf{Generate exactly} \texttt{\{n\_candidates\}} \textbf{DIFFERENT corrective trajectories in ONE JSON array.} Each object must contain these string fields:
\begin{enumerate}[leftmargin=*, itemsep=0.5pt, parsep=0pt, topsep=1pt, partopsep=0pt]
\item \textbf{error\_step:} a concise first-person diagnosis naming where and why the first error occurred. With multiple trajectories, describe the common pattern of confusion; with one, describe its concrete first error.
\item \textbf{preserved\_prefix:} with one trajectory, its LONGEST correct prefix, copied VERBATIM from the beginning up to but excluding the first erroneous step; with multiple trajectories, any shared correct prefix; empty if none.
\item \textbf{correction:} continue from \texttt{preserved\_prefix} to a complete correct solution. Match the student's vocabulary, formatting, and tone. End with exactly "The answer is: [value]".
\item \textbf{final\_answer:} only the final numerical or symbolic answer.
\end{enumerate}
\par
The correction is intentionally more structured than the student's attempts because the student needs scaffolding, but still matches the student's vocabulary and tone.
\textbf{Localize the first error} and diversify the \texttt{\{n\_candidates\}} continuations jointly in this single call. \textbf{Never invent an error in correct reasoning.} Avoid apologies, meta-commentary such as "Let me reconsider", and explanations outside the array.
\par
\textbf{Reference output-format examples:} \texttt{\{trajectory\_examples\}}

\textbf{Task data:} \texttt{\{task\_data\}}
\par\endgroup
\end{tcolorbox}
\par

\par
\subsection{Prompt for Sporadic Errors}\label{appendix:prompt_sporadic}

For sporadic errors, the student answers most attempts
correctly and only occasionally fails. The student already
possesses the relevant skills, so the prompt instructs the
teacher to preserve the longest possible correct prefix $p$
and apply the smallest possible edit at the divergence point.

\begin{tcolorbox}[breakable, lines before break=4, boxsep=2pt, left=5pt, right=5pt, top=2pt, bottom=2pt, before skip=4pt, after skip=4pt, colback=gray!5, colframe=black!60, title={\textbf{Prompt Template for Sporadic Errors}}, fonttitle=\small]
\begingroup\small
\setlength{\parskip}{1pt}
You are an expert Math Corrector. You specialize in MINIMAL, PRECISE corrections.
\par
\textbf{The student usually answers correctly.} Preserve as much of the reasoning as possible and change only the localized error. \textbf{Vary the local repair:} a direct fix, brief inline verification, a slightly different computation avoiding the same error, or further minimal variations. \textbf{Do not restructure correct steps.}
The localized slip may be an arithmetic mistake, sign error, misapplied formula, or off-by-one in an otherwise reasonable solution. The overall problem-solving approach and structure are sound; only the erroneous step needs to be repaired.
\par
\textbf{Local-repair examples:} trajectory~1 directly replaces the wrong step with the correct computation; trajectory~2 corrects with a brief inline verification; trajectory~3 takes a slightly different computational path that avoids the specific error; further trajectories explore additional minimal variations.

\par\noindent \textbf{Generate exactly} \texttt{\{n\_candidates\}} \textbf{DIFFERENT corrective trajectories in ONE JSON array.} Each object must contain these string fields:
\begin{enumerate}[leftmargin=*, itemsep=0.5pt, parsep=0pt, topsep=1pt, partopsep=0pt]
\item \textbf{error\_step:} a concise first-person diagnosis naming where and why the first error occurred. Describe the specific localized slip, for example "I made an arithmetic error when adding\ldots".
\item \textbf{preserved\_prefix:} the student's LONGEST correct prefix, copied VERBATIM from the beginning up to but excluding the first erroneous step; empty if none.
\item \textbf{correction:} continue from \texttt{preserved\_prefix} to a complete correct solution. Match the student's vocabulary, formatting, and tone. End with exactly "The answer is: [value]".
\item \textbf{final\_answer:} only the final numerical or symbolic answer.
\end{enumerate}
\par
The correction should feel like a seamless continuation of the student's own work. Keep the correction short.
\textbf{Localize the first error} and diversify the \texttt{\{n\_candidates\}} continuations jointly in this single call. \textbf{Never invent an error in correct reasoning.} Avoid apologies, meta-commentary such as "I made a mistake here" or "Let me reconsider", and explanations outside the array.
\par
\textbf{Reference output-format examples:} \texttt{\{trajectory\_examples\}}

\textbf{Task data:} \texttt{\{task\_data\}}
\par\endgroup
\end{tcolorbox}
\par

\par
\subsection{Uniform Minimal-Correction Prompt}\label{appendix:prompt_uniform}

The ablation without profiling uses the same
minimal-correction policy for every error, instead of the
three mode-specific prompts.
The uniform prompt receives only the problem, representative incorrect
trace, and correct answer, without profiling metadata.

\begin{tcolorbox}[breakable, lines before break=4, boxsep=2pt, left=5pt, right=5pt, top=2pt, bottom=2pt, before skip=4pt, after skip=4pt, colback=gray!5, colframe=black!60, title=\textbf{Uniform Minimal-Correction Prompt}, fonttitle=\small]
{\small
\setlength{\parskip}{1pt}
\textbf{[System]} You are an expert Math Corrector
specializing in minimal, precise corrections.

\textbf{Context.}
The student attempted this problem and made a mistake in
an otherwise reasonable solution. Preserve as much of the
student's original reasoning as possible and fix only
what is broken.

\textbf{Your Task.}
Generate exactly \texttt{\{n\_candidates\}} DIFFERENT
corrective trajectories. Each trajectory must:
\begin{enumerate}[leftmargin=*, itemsep=0.5pt, parsep=0pt, topsep=1pt, partopsep=0pt]
\item Preserve the maximum correct prefix: copy the student's
reasoning VERBATIM up to the point of error, the longest
correct prefix possible.
\item Pinpoint the exact error: identify the specific
calculation, substitution, or logical misstep in a
first-person \texttt{error\_step}, rather than stating
only \textquotedblleft I was wrong\textquotedblright.
\item Apply a minimal correction: fix ONLY what is wrong;
do not restructure or rewrite correct steps.
\item Vary the correction style across the
\texttt{\{n\_candidates\}} trajectories.
\end{enumerate}

\par\noindent\begin{minipage}{\linewidth}
\textbf{Style Requirements.}
\begin{itemize}[leftmargin=*, itemsep=0.5pt, parsep=0pt, topsep=1pt, partopsep=0pt]
\item Preserve the student's writing style, formatting, and tone.
\item No meta-commentary such as
\textquotedblleft I made a mistake here\textquotedblright{}
or \textquotedblleft Let me reconsider\textquotedblright.
\item Every trajectory MUST end with exactly:
\textquotedblleft The answer is: [value]\textquotedblright.
\end{itemize}
\end{minipage}\par

\textbf{Output Format.}
A JSON array of \texttt{\{n\_candidates\}} objects with fields
\{\texttt{error\_step}, \texttt{preserved\_prefix},
\texttt{correction}, \texttt{final\_answer}\}.

\textbf{Reference Examples.}
\texttt{\{trajectory\_examples\}}

\textbf{Task Data.}
\texttt{\{problem\}} / \texttt{\{student\_reasoning\}} /
\texttt{\{correct\_answer\}}
}
\end{tcolorbox}
\par

\section{Error Mode Distribution}\label{appendix:error_mode_distribution}

\begingroup
Figure~\ref{fig:error_mode_pie} shows the distribution of diagnosed error modes among 63,062 problems before teacher-candidate correctness filtering in the Llama-3.1-8B first construction pass. Of these problems, 62,740 are retained after filtering to form the first-pass training corpus used in the ablation and window-size studies.

\begin{figure}[H]
\centering
\includegraphics[width=0.55\columnwidth]{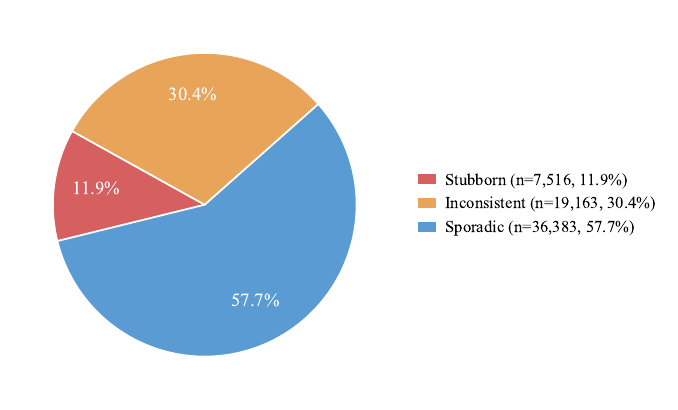}
\caption{{Distribution of diagnosed error modes among 63,062 problems before teacher-candidate correctness filtering in the Llama-3.1-8B first construction pass.}}
\label{fig:error_mode_pie}
\end{figure}
\par\endgroup

\section{Profiling Threshold Sensitivity}\label{appendix:threshold}

We evaluate threshold sensitivity on the same 63,062 problems
from the Llama-3.1-8B first construction pass shown in
Figure~\ref{fig:error_mode_pie}, before teacher-candidate correctness
filtering. The problem set and the five sampled student responses
per problem are held fixed across threshold settings.

Our behavioral profiling classifies each problem based on
two thresholds: the error rate threshold $\alpha_t$ and the
consistency threshold $\kappa_t$. With $K\!=\!5$ samples per
problem, $\alpha$ takes values in
$\{0, 0.2, 0.4, 0.6, 0.8, 1.0\}$ and $\kappa$ takes a
similarly discrete set of values determined by the number of
incorrect samples. This discretization means that the
classification is invariant to any threshold perturbation
that stays within the same discrete interval, so several
threshold pairs reproduce the default classification exactly.

\begin{figure}[H]
\centering
\includegraphics[width=0.7\columnwidth]{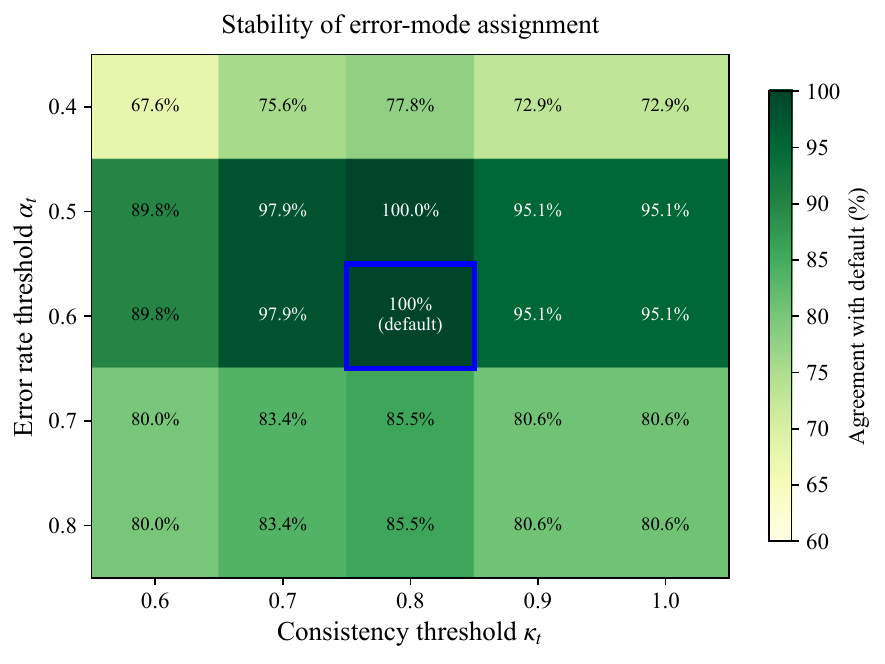}
\caption{Agreement (\%) with the default thresholds
$(\alpha_t, \kappa_t) = (0.6, 0.8)$ on the same 63,062-problem set as
Figure~\ref{fig:error_mode_pie}. Each cell shows the
fraction of problems whose error mode matches the
classification under the default setting, with the default
cell at 100\%. Cells adjacent to the default exhibit high
agreement due to the discrete nature of $\alpha$ and
$\kappa$ under $K\!=\!5$ sampling.}
\label{fig:threshold}
\end{figure}

Figure~\ref{fig:threshold} reports the agreement with the
default thresholds $(\alpha_t, \kappa_t) = (0.6, 0.8)$,
measured as the fraction of problems whose diagnosed mode
matches the default. The neighboring setting
$(\alpha_t, \kappa_t) = (0.5, 0.8)$ reproduces the default
classification exactly (100\%), and all cells in
$\alpha_t \in \{0.5, 0.6\}$ with $\kappa_t \in \{0.7, 0.9, 1.0\}$
remain at 95.1\%--97.9\% agreement. Agreement degrades smoothly
as the thresholds move further from the default: lowering
$\alpha_t$ to 0.4 drops agreement to 67.6\%--77.8\%, because
problems with $\alpha\!=\!0.4$ are reclassified from sporadic
to stubborn or inconsistent, while raising $\alpha_t$ to
0.7--0.8 yields 80.0\%--85.5\%. These represent substantial
shifts in the classification criteria rather than fine
perturbations. In practice, the discrete structure of
$\alpha$ and $\kappa$ under small $K$ provides a natural
robustness to threshold selection: the method does not
require fine-grained tuning to achieve stable classifications.

\begin{figure}[H]
\centering
\includegraphics[width=0.85\columnwidth]{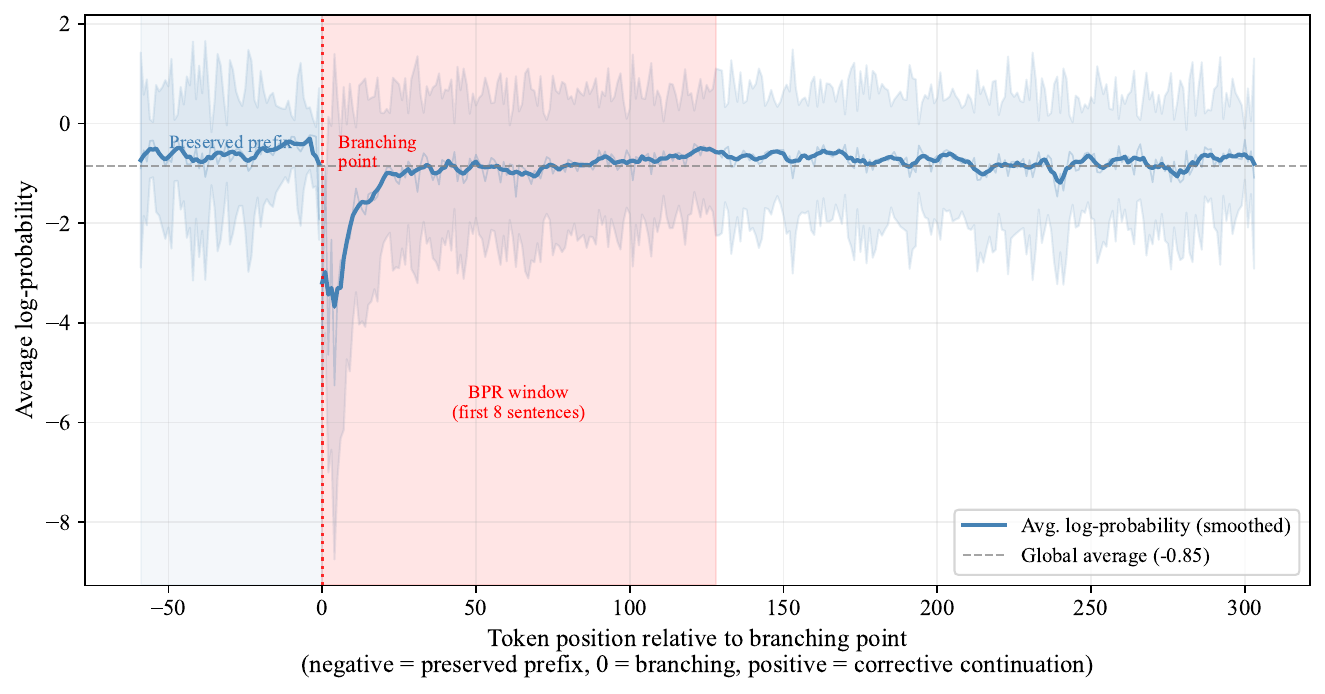}
\caption{Average per-token log-probability of the
Llama-3.1-8B student against token position relative to
the branching point, over 200 examples. Both the
preserved prefix and the recovered tail of the corrective
continuation sit near the global average of $-0.85$, while
only a narrow trough around the branching point falls to
about $-3.5$. The shaded red region marks the BPR scoring
window covering the first eight sentences.}
\label{fig:logprob_curve}
\end{figure}

\raggedbottom
\begin{samepage}
\section{Token-Level Log-Probability Curve}\label{appendix:logprob}

To verify the motivation behind branching point reachability scoring, we plot the per-token log-probability of the student against position relative to the branching point, averaged over 200 training examples. Figure~\ref{fig:logprob_curve} shows the preserved prefix sitting near the global average of $-0.85$, dropping sharply to about $-3.5$ at the branching point, and recovering within roughly 25 tokens. Global scoring averages over the long high-probability tail and washes out the few low-probability tokens that distinguish reachable from unreachable continuations, whereas restricting evaluation to the shaded BPR window concentrates the signal on the positions where candidates diverge, {matching the 1.7\% gain over global scoring in both ablation studies.}

\par\end{samepage}

\begingroup
\section{Additional Experimental Results}\label{appendix:additional_results}

This section provides the complete supplementary comparisons
and analyses summarized in Section~\ref{sec:experiments}.

\subsection{Shared-Source Coverage Control}\label{appendix:coverage_results}

We map Ours and LEMMA to their original GSM8K and MATH
problems and restrict both methods to the same 12,217 shared
source problems, retaining one training example per source
problem for each method. This controls source-problem
coverage and example count while retaining each method's
supervision format. Table~\ref{tab:shared_source_control}
reports the DeepSeek-Math-7B results. Ours achieves 77.1\%
versus LEMMA's 75.7\%, with higher OOD results while LEMMA
remains stronger on MATH. This is a coverage control,
rather than a causal test of BPR alone.

\begin{table}[H]

\centering
\caption{Accuracy (\%) on DeepSeek-Math-7B after restricting each method to the same 12,217 source problems, with one training example per source problem. This controls source coverage and sample count while retaining each method's supervision format.}
\label{tab:shared_source_control}
\small
\setlength{\tabcolsep}{4pt}
\begin{tabular*}{\linewidth}{@{\extracolsep{\fill}}lrrrrrr}
\toprule
\textbf{Method} & \textbf{GSM8K} & \textbf{MATH} & \textbf{SVAMP} & \textbf{ASDiv} & \textbf{MAWPS} & \textbf{Avg.} \\
\midrule
Ours & \textbf{80.3} & 45.2 & \textbf{81.4} & \textbf{87.3} & \textbf{91.2} & \textbf{77.1} \\
LEMMA & 77.5 & \textbf{47.7} & 78.1 & 86.4 & 88.6 & 75.7 \\
\bottomrule
\end{tabular*}
\end{table}

\subsection{First-Divergence Localization}\label{appendix:localization_results}

\paragraph{Boundary localization.}
A manual audit
of 50 randomly sampled traces from the
65.76K DeepSeek configuration finds that 44 boundaries are accurate,
four are too early, and two are too late. Keeping teacher candidate
pools fixed, we perturb BPR boundaries on 12\% of questions
with a preserved prefix: 8\% move one reasoning step earlier
and 4\% one step later. We recompute scores under the shifted
prefixes and windows, reselect trajectories, and retrain.
Table~\ref{tab:boundary_perturbation} reports this single
run, in which average accuracy changes from 80.0\%
to 79.7\%. This checks the sensitivity of BPR scoring and
candidate reselection to boundary localization.

\begin{table}[H]

\centering
\caption{Accuracy (\%) after boundary perturbation for the 65.76K DeepSeek-Math-7B configuration. Teacher candidate pools are fixed; BPR is recomputed under shifted boundaries before trajectory reselection and retraining. This is a single-run selection-sensitivity experiment.}
\label{tab:boundary_perturbation}
\small
\setlength{\tabcolsep}{4pt}
\begin{tabular*}{\linewidth}{@{\extracolsep{\fill}}lrrrrrr}
\toprule
\textbf{Setting} & \textbf{GSM8K} & \textbf{MATH} & \textbf{SVAMP} & \textbf{ASDiv} & \textbf{MAWPS} & \textbf{Avg.} \\
\midrule
Original & 83.6 & \textbf{47.1} & \textbf{83.9} & \textbf{90.1} & \textbf{95.3} & \textbf{80.0} \\
Perturbed & \textbf{83.8} & 46.6 & 83.6 & 89.4 & 94.9 & 79.7 \\
\bottomrule
\end{tabular*}
\end{table}

{\subsection{{Teacher Variation}}\label{appendix:robustness_results}}

{Table~\ref{tab:teacher_variation} compares teachers in single SFT runs using the same fixed seed. Replacing DeepSeek-V3.2 with DeepSeek-V4-Flash gives 81.0\% average accuracy, versus 80.6\%, with all other settings held fixed.}

\begin{table}[H]

\centering
\caption{Accuracy (\%) with different teachers for DeepSeek-Math-7B using 88.90K constructed examples. Both rows report single SFT runs using the same fixed seed. All settings other than the teacher are held fixed.}
\label{tab:teacher_variation}
\small
\setlength{\tabcolsep}{4pt}
\begin{tabular*}{\linewidth}{@{\extracolsep{\fill}}lrrrrrr}
\toprule
\textbf{Teacher} & \textbf{GSM8K} & \textbf{MATH} & \textbf{SVAMP} & \textbf{ASDiv} & \textbf{MAWPS} & \textbf{Avg.} \\
\midrule
DeepSeek-V3.2 & 84.1 & 48.6 & 85.3 & 90.1 & \textbf{95.0} & 80.6 \\
DeepSeek-V4-Flash & \textbf{84.7} & \textbf{49.1} & \textbf{85.7} & \textbf{90.8} & 94.9 & \textbf{81.0} \\
\bottomrule
\end{tabular*}
\end{table}

\subsection{Inference Behavior}\label{appendix:inference_results}

Table~\ref{tab:inference_reflection_audit} reports an audit
of 300 test outputs per benchmark for each 88.90K
student, totaling 1,500 outputs per student. An LLM judge
labels explicit revision of an earlier step as REFLECTION
and other solutions as DIRECT. DIRECT outputs account for
90.5\% and 95.3\% of Llama and DeepSeek outputs, with
accuracies of 81.9\% and 82.1\%; REFLECTION accuracies are
33.1\% and 35.7\%. The share-weighted accuracies are 77.3\%
and 79.9\%. This provides indirect evidence that corrective
training is compatible with question-only inference without
routinely triggering explicit revision.

\begin{table}[H]

\centering
\caption{Test-output audit for the 88.90K students: 300 outputs per benchmark, totaling 1,500 per student. REFLECTION denotes explicit revision within a test-time solution; DIRECT denotes no such revision. Shares and accuracies are percentages.}
\label{tab:inference_reflection_audit}
\small
\setlength{\tabcolsep}{4pt}
\begin{tabular*}{\linewidth}{@{\extracolsep{\fill}}lrrrr}
\toprule
& \multicolumn{2}{c}{\textbf{DIRECT}} & \multicolumn{2}{c}{\textbf{REFLECTION}} \\
\cmidrule(lr){2-3}\cmidrule(lr){4-5}
\textbf{Student} & \textbf{Share} & \textbf{Accuracy} & \textbf{Share} & \textbf{Accuracy} \\
\midrule
Llama-3.1-8B & 90.5 & \textbf{81.9} & 9.5 & 33.1 \\
DeepSeek-Math-7B & 95.3 & \textbf{82.1} & 4.7 & 35.7 \\
\bottomrule
\end{tabular*}
\end{table}

\par\endgroup

\raggedbottom
\section{Case Examples of Token-Level Log-Probability}\label{appendix:logprob_cases}

Beyond the average curve in Figure~\ref{fig:logprob_curve}, we present three
individual training examples in Figures~\ref{fig:logprob_case_1},
\ref{fig:logprob_case_2}, and \ref{fig:logprob_case_3} that illustrate the same
pattern at the level of single problems. Each figure displays the problem
statement, the preserved prefix, the BPR window covering the first eight
sentences after the branching point, and a portion of the later continuation,
alongside the corresponding per-token log-probability curve under the
Llama-3.1-8B student.

In all three cases, the preserved prefix and the later continuation sit close
to the global average, while the BPR window concentrates a sequence of sharp
dips that pull its mean roughly 1.2 to 1.7 nats below the tail average. This gap is consistent with the averaged behavior in Figure~\ref{fig:logprob_curve} and illustrates the concentration of low-probability tokens near the branching point. The first example introduces an equation-based framework that
identifies given quantities, defines variables, and sets up earnings
equations for a marble pricing word problem, the second rewrites both sides
to a common base before equating exponents on an exponential equation, and
the third lays out grouped variables for hourly wage, working days, and
total hours on a monthly income word problem. These examples illustrate the probability cost of the initial corrective steps. The performance benefit of using $L=8$ is evaluated in the component ablation and window-size studies in Section~\ref{sec:ablation}.

\clearpage
\noindent
\begin{minipage}{\linewidth}
\begin{tcolorbox}[ colback=gray!5, colframe=black!60, title=\textbf{Case Example 1: Marble Pricing Word Problem}, fonttitle=\small]
{\small
\textbf{Problem.} Alvin is selling his marble set. He has 100 marbles. 20\% are white, 30\% are black, and the rest are different colors. He sells the white ones for a nickel each, the black ones for a dime each, and the colors for \$x each. How much does he earn? If we know the answer to the above question is 14, what is the value of unknown variable x?

\textbf{Preserved prefix.} Alvin has 100 marbles. 20\% are white: 0.20 x 100 = 20 white marbles 30\% are black: 0.30 x 100 = 30 black marbles The rest are different colors: 100 - 20 - 30 = 50 different colored marbles

\textbf{BPR window.} Let's solve this using a clear equation-based framework, sticking to dollars to avoid unit confusion. Step 1: Identify what's given and what's asked. Given: Total marbles = 100. Percentages and prices for white and black marbles. Total earnings = \$14. Need to find x (price per colored marble in dollars).

\textbf{Later continuation.} Step 2: Define variables. Let x = price in dollars for one colored marble. Step 3: Set up equations for earnings from each type. Earnings from white marbles = 20 marbles * \$0.05 = \$1 Earnings from black marbles = 30 marbles * \$0.10 = \$3 Earnings from colored marbles = 50 marbles * x dollars = 50x dollars Step 4: Write the total earnings equation. Total earnings = \$1 + \$3 + 50x = \$14 Step 5: Solve for x. Combine constants: \$4 + 50x = \$14 Subtract \$4: 50x = \$10 Divide by 50: x = \$10 / 50 = \$0.20 Step 6: Verify. Check: White: 20*0.05=1, Black: 30*0.10=3, Colored: 50*0.20=10. Sum = 1+3+10=14. Correct. The answer is: 0.20
}
\end{tcolorbox}

\begin{figure}[H]
\centering
\includegraphics[width=\columnwidth]{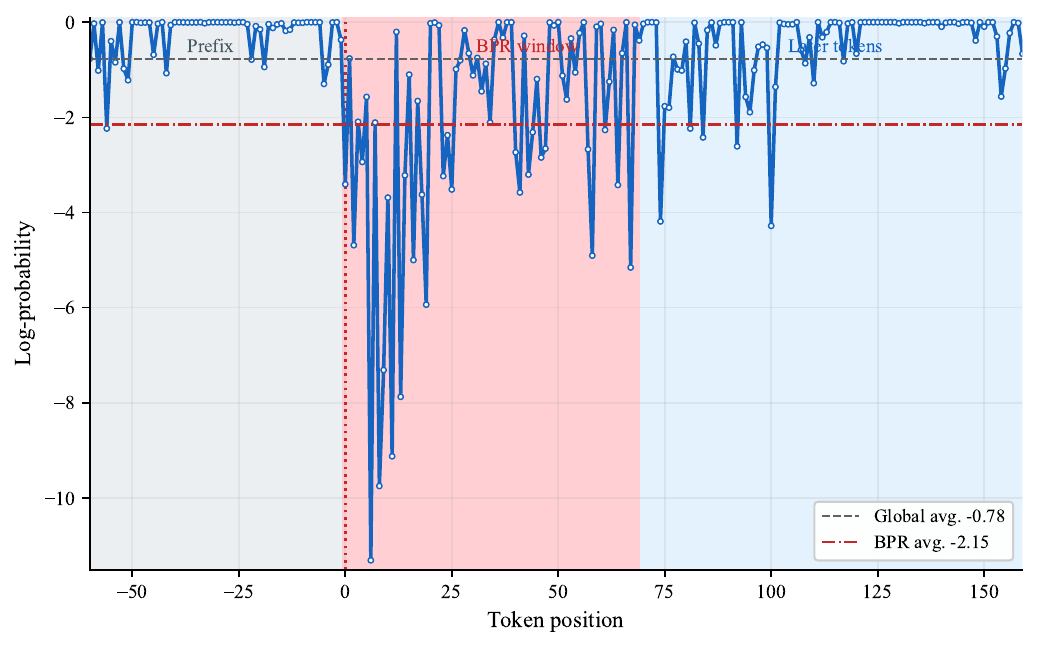}
\caption{Per-token log-probability for an individual training example on a
marble pricing word problem. The preserved prefix and later continuation track
the global average, while the BPR window absorbs sharp dips that lower its
mean to $-2.15$ against a tail average of $-0.45$, yielding a gap of
$1.70$ nats.}
\label{fig:logprob_case_1}
\end{figure}
\end{minipage}
\par\medskip

\clearpage
\noindent
\begin{minipage}{\linewidth}
\begin{tcolorbox}[ colback=gray!5, colframe=black!60, title=\textbf{Case Example 2: Exponential Equation}, fonttitle=\small]
{\small
\textbf{Problem.} Solve for $x$, if $8^{4x-6}=\left(\frac{1}{2}\right)^{x+5}$.

\textbf{Preserved prefix.} To solve for $x$, we first need to express both sides of the equation with the same base. We can rewrite $8$ as $2^3$ and $\frac{1}{2}$ as $2^{-1}$.

\textbf{BPR window.} Let's follow a clear, equation-based framework. Step 1 -- Identify the goal: Solve for x in $8^{4x-6}=\left(\frac{1}{2}\right)^{x+5}$. Step 2 -- Choose a common base. The easiest is base 2 because 8 = 2\textasciicircum{}3 and 1/2 = 2\textasciicircum{}\{-1\}. Step 3 -- Rewrite each side using base 2.

\textbf{Later continuation.} Left side: $8^{4x-6} = (2^3)^{4x-6} = 2^{3(4x-6)} = 2^{12x-18}$. Right side: $\left(\frac{1}{2}\right)^{x+5} = (2^{-1})^{x+5} = 2^{-1(x+5)} = 2^{-x-5}$. Step 4 -- Set the exponents equal since the bases are the same and the equation is an equality. $12x - 18 = -x - 5$ Step 5 -- Solve the linear equation systematically. Add x to both sides: $12x + x - 18 = -5$ $\rightarrow$ $13x - 18 = -5$. Add 18 to both sides: $13x = 13$. Divide both sides by 13: $x = 1$. Step 6 -- Quick check: Plug x=1 back into the original equation. Left: $8^{4(1)-6} = 8^{-2} = 1/64$. Right: $(1/2)^{1+5} = (1/2)^6 = 1/64$. It matches. The answer is: 1
}
\end{tcolorbox}

\begin{figure}[H]
\centering
\includegraphics[width=\columnwidth]{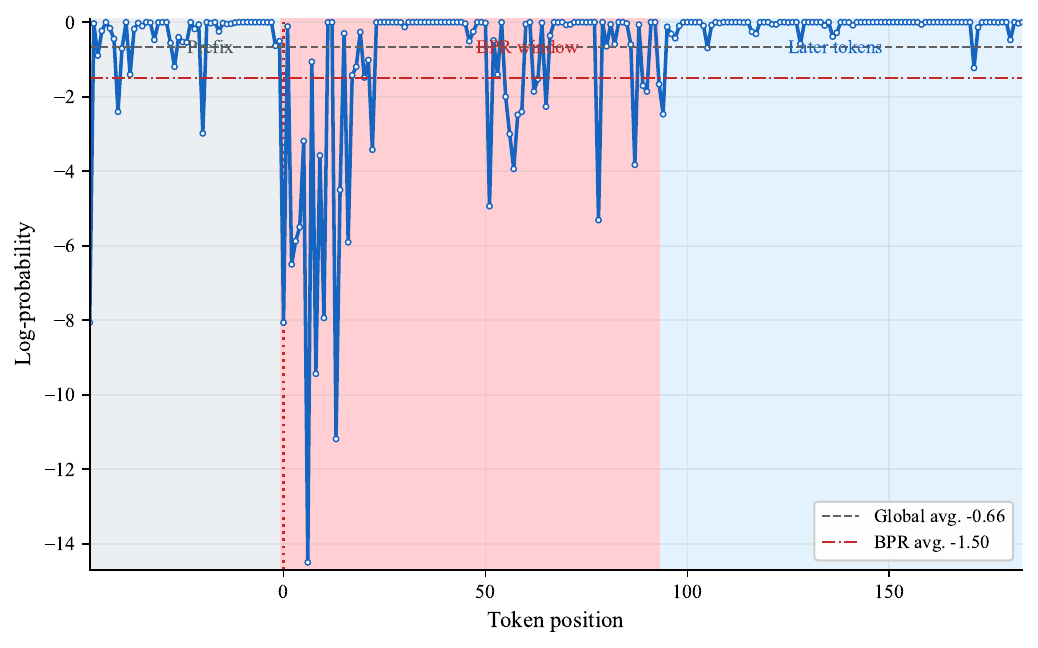}
\caption{Per-token log-probability for an exponential equation. The BPR
window contains the framework setup tokens that rewrite both sides to a
common base before equating exponents, lowering the BPR mean to $-1.50$
against a tail average of $-0.09$, with a gap of $1.41$ nats.}
\label{fig:logprob_case_2}
\end{figure}
\end{minipage}
\par\medskip

\clearpage
\noindent
\begin{minipage}{\linewidth}
\begin{tcolorbox}[ colback=gray!5, colframe=black!60, title=\textbf{Case Example 3: Monthly Income Word Problem}, fonttitle=\small]
{\small
\textbf{Problem.} If John works for 12 hours every other day and receives a 30\% raise from his previous \$20 per hour job, what is his monthly income for a 30-day month?

\textbf{Preserved prefix.} First, let's calculate John's new hourly wage: \$20 * 1.30 = \$26 per hour.

\textbf{BPR window.} Step 1: Understand the quantities given. Previous wage: \$20 per hour. Raise: 30\%. Hours per workday: 12 hours. Work schedule: every other day. Month length: 30 days. We are asked: monthly income. Step 2: Set up variables and equations.

\textbf{Later continuation.} Let new\_wage = old\_wage * (1 + raise\_percent/100) = 20 * 1.30 = \$26. Let work\_days = total\_days / 2 = 30 / 2 = 15 days. Let total\_hours = work\_days * hours\_per\_day = 15 * 12 = 180 hours. Let monthly\_income = new\_wage * total\_hours = 26 * 180. Step 3: Execute computation step by step. 26 * 180 = 26 * (100 + 80) = (26*100) + (26*80) = 2600 + 2080 = 4680. Step 4: Verify. Half of 30 days is 15 days, which is reasonable. 15 days * 12 hours = 180 hours. \$26 per hour * 180 hours = \$4680, which is a plausible monthly income. The answer is: 4680
}
\end{tcolorbox}

\begin{figure}[H]
\centering
\includegraphics[width=\columnwidth]{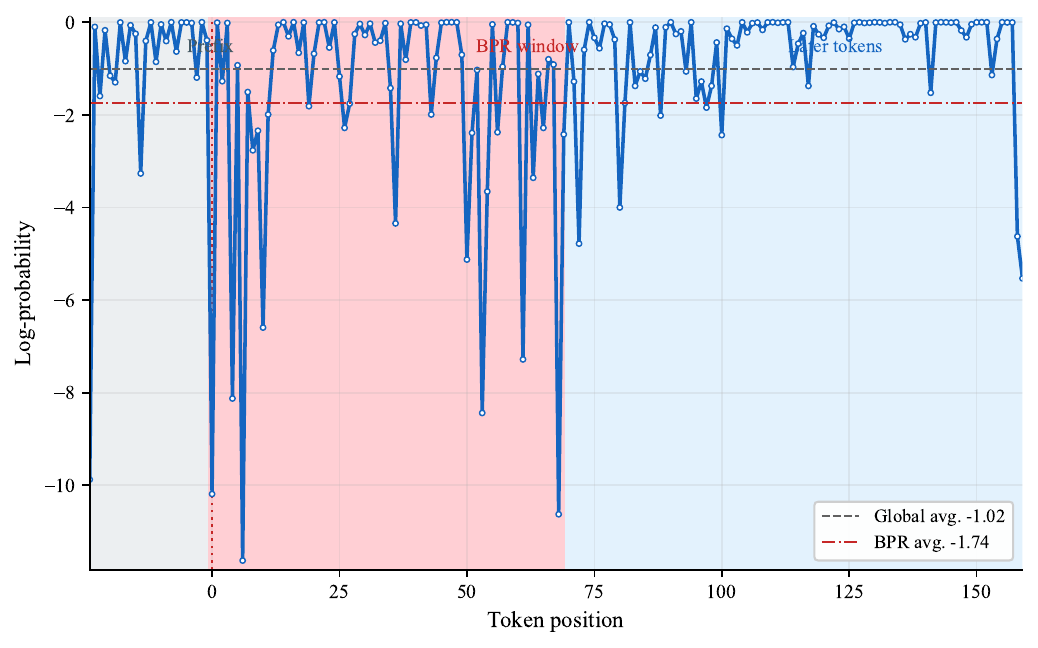}
\caption{Per-token log-probability for a monthly income word problem. The BPR window summarizes the given quantities and introduces the variable-setting stage, with a mean of $-1.74$ against a tail average of $-0.57$, yielding a gap of $1.17$ nats.}
\label{fig:logprob_case_3}
\end{figure}
\end{minipage}
\par\medskip

\flushbottom

\end{document}